\documentclass{article}

\usepackage[preprint]{corl_2026} 

\usepackage{amsmath,amsfonts,bm}

\def\eqref#1{equation~\ref{#1}}

\def\1{\bm{1}}

\DeclareMathAlphabet{\mathsfit}{\encodingdefault}{\sfdefault}{m}{sl}
\SetMathAlphabet{\mathsfit}{bold}{\encodingdefault}{\sfdefault}{bx}{n}

\usepackage{wrapfig}
\usepackage{hyperref}
\usepackage{url}
\usepackage{multirow}
\usepackage{makecell}
\usepackage{graphicx}
\usepackage{epsfig} 
\usepackage{mathptmx}
\usepackage{amsmath} 

\usepackage{amssymb}  
\usepackage{amsthm}
\usepackage{wasysym}
\usepackage{mathrsfs}
\usepackage[mathcal]{euscript}
\usepackage{bbm}
\usepackage{lipsum}
\usepackage{algorithm}
\usepackage{algpseudocode}
\usepackage{tcolorbox}
\tcbuselibrary{breakable}
\usepackage[normalem]{ulem}
\usepackage{tabularx}
\usepackage{booktabs}
\usepackage{array}
\usepackage{enumitem}

\usepackage{subcaption}
\usepackage{graphicx}
\usepackage{minted}

\title{Understanding and Mitigating the Video-Action Generalization Gap via Temporal Ratio}

\author{%
    Utkarsh A. Mishra\textsuperscript{1*}, 
 Yongxin Chen\textsuperscript{1},    Danfei Xu\textsuperscript{1}, {Yang Liu\textsuperscript{2}, 
     Xi Chen\textsuperscript{2}}, Jiayuan Mao\textsuperscript{2}
    \\
    \textsuperscript{1}Georgia Tech,
    \textsuperscript{2}Amazon FAR \\ \textsuperscript{*}Work done during an internship at Amazon FAR
}

\begin{document}
\maketitle

\begin{abstract}
Generative video foundation models exhibit strong compositional priors, yet world-action models (WAMs) and video-action models (VAMs) often lose these priors after finetuning on robotic action data. We refer to this discrepancy as the {\bf video-action generalization gap}. In this paper, we systematically investigate this gap by evaluating a comprehensive design space of VAMs, demonstrating that standard design choices yield no emergent explanation pattern. To explain this behavior, we introduce the Temporal Ratio (TR), an attention-based measure of how strongly the action head relies on future latent rollouts relative to the anchored current frame. TR has two key properties: first, a model's structural reliance on future-predictive latents, measured via TR, acts as a predictor of its compositional generalization capacity; second, it natively fluctuates based on task phase, shifting attention to future frames during planning and reverting to the present frame for precise manipulation. Finally, based on these findings, we propose an inference-time adaptive guidance method, which exploits this intrinsic feature attention pattern to dynamically amplify compositional video conditioning signals precisely when the policy relies on future rollouts. Evaluated on the LIBERO benchmark and real-world tasks, our approach mitigates the OOD-ID compositional generalization gap. More details: \url{https://umishra.me/temporal-ratio/}
\end{abstract}



\section{Introduction}
\label{sec:intro}

Compositional generalization~\cite{farid2025drives, chen2025robohiman, zhou2025libero} is essential for autonomous manipulation: a robot should recombine seen objects, receptacles, and primitives into unseen task sequences. Existing policies often fail under such recompositions because they overfit to visual correlations such as object coordinates~\cite{li2025task, fang2026vision, fei2025libero, zhou2025libero}. For example, a policy trained on ``put the cream cheese in the bowl'' and ``put the bowl on top of the cabinet'' may still fail on ``put the cream cheese on top of the cabinet,'' despite involving only previously seen objects, receptacles, and primitives. An emerging direction is to use generative video foundation models~(VFMs)~\cite{ali2025world, wan2025wan, VeoTeam2025} for action learning, leveraging video models trained on internet-scale video data with strong compositional generalization by capturing temporal coherence, object interactions, and physical plausibility.

\begin{wrapfigure}{r}{0.43\textwidth}
\vspace{-1em}
  \begin{center}
    \includegraphics[trim={0 0.2cm 0 0},clip,width=\linewidth]{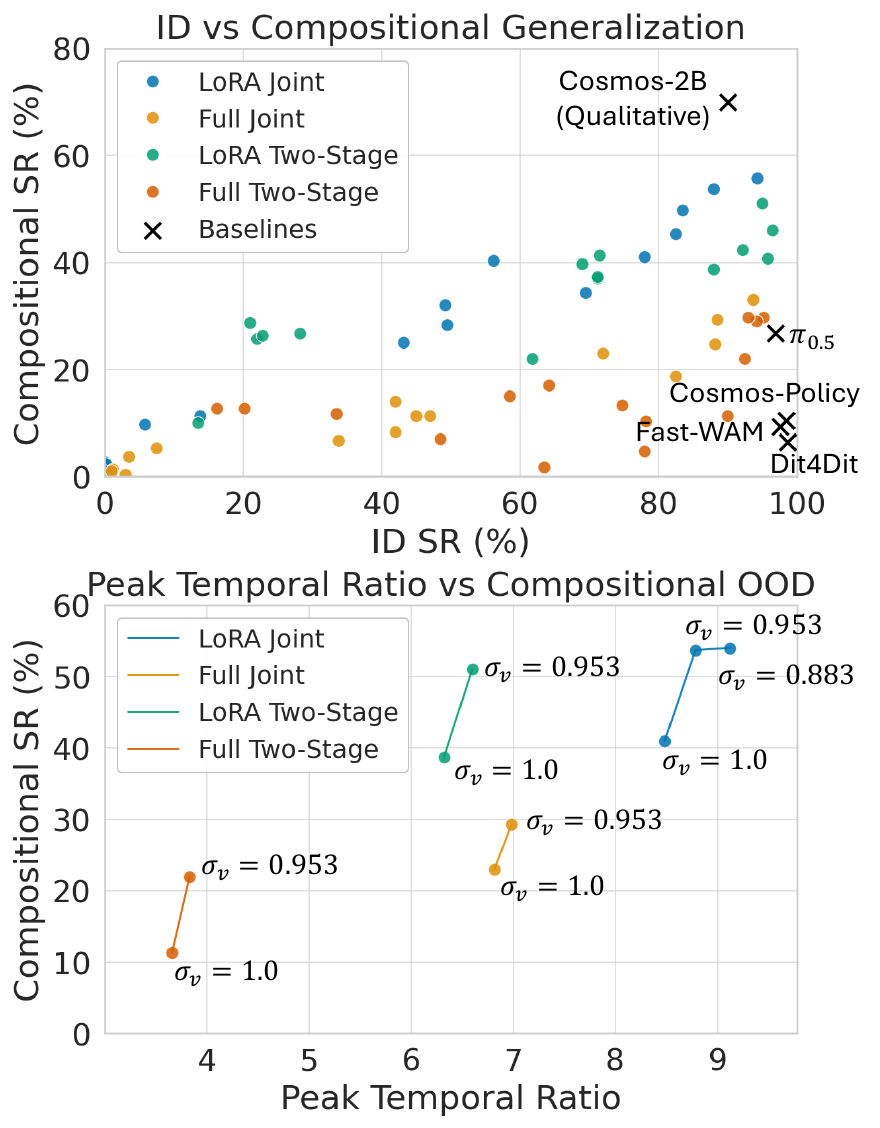}
  \end{center}
  \caption{\textbf{Interpretable metric to explain performance gap between ID and compositional OOD.} (Top) We study the design space of VAMs over video feature-extraction noise level $\sigma_v$ and other design choices, finding a fragmented ID-OOD performance gap without a clear consensus. (Bottom) We propose Temporal Ratio (TR) as a diagnostic tool to interpret OOD performance and eventually improve it.}
  \label{fig:id_ood_tr_pattern}
  \vspace{-1.5em}
\end{wrapfigure}

Consequently, the key challenge is how to transfer these pretrained video priors into robotic control. Current World-Action Models (WAMs)~\cite{kim2026cosmos, ye2026world} and Video-Action Models (VAMs)~\cite{pai2025mimic, ma2026dit4dit, yuan2026fast} typically either (1) predict future video~\cite{li2025novaflow, du2024video} (or corresponding latent features~\cite{pai2025mimic, ma2026dit4dit}) and infer actions through inverse dynamics, or (2) jointly model video and action tokens within a unified transformer~\cite{kim2026cosmos, ye2026world, li2026causal, liang2025video, li2025unified, zhu2025unified}. However, despite strong in-domain performance, these methods consistently fail to preserve the compositional generalization abilities of their underlying video backbones~(\autoref{fig:id_ood_tr_pattern}, Top) after finetuning on action data~\cite{feng2026harmowam}.
We refer to this discrepancy as the \textbf{video-action generalization~(VAG) gap}. This motivates the following research question: what is the 
right interface between VFMs and action prediction that preserves compositional priors?

In this work, we examine a class of VAMs that use latent features from pre-trained video foundation models and learn a latent inverse-dynamics action head~\cite{pai2025mimic, ma2026dit4dit, yuan2026fast}. Such models preserve the video model's generative prior by feeding latent video features to an action head trained with flow matching~\cite{lipman2022flow}. 
To understand the properties of this interface, we systematically study its design space across backbone adaptation and video-latent feature extraction choices. This reveals a fragmented manifold of the in-distribution~(ID) and out-of-distribution~(OOD) performance gap~(\autoref{fig:id_ood_tr_pattern}, Top), motivating a more mechanistic diagnostic.


Our key observation is that compositional generalization depends on whether the action head actually uses the predicted future rollout. We introduce the \textbf{Temporal Ratio} (TR), the attention mass assigned by the action head to future latent frames relative to the anchored current frame. Illustrated in \autoref{fig:id_ood_tr_pattern} (Bottom): TR is strongly correlated with a model's underlying generalization capacity. While success rates look like noise, TR acts as a direct predictor of the ID-OOD performance gap. This gives us an opportunity to close the ID-OOD performance gap by treating TR as a runtime signal to implement inference-time guidance. We also observe that TR varies with task phase: it rises during ``planning phase'' (e.g., what object to pick, where to place) and drops during precise local manipulation (e.g., grasping or placement). We exploit this intrinsic property via \emph{TR-Adaptive Guidance}, which amplifies language and plan conditioning when the policy enters a planning regime and relaxes guidance during precise manipulation. We evaluate on LIBERO compositional tasks~\cite{Liu2023LIBEROBK, li2025task} and real-world bimanual tasks, improving OOD success while preserving ID performance.
\vspace{-0.5em}
\section{Design Space of Video-Action Models}
\label{sec:design-space}
\vspace{-0.5em}

We study a particular class of VAMs similar to the design of~\cite{pai2025mimic, ma2026dit4dit}. As shown in~\autoref{fig:architecture}, the architecture combines (1) video denoising via flow matching with partially denoised video feature extraction and (2) an action head that denoises actions from the video features from (1). We explore this design space to understand how a video backbone facilitates action generation.

\vspace{-0.5em}
\subsection{Background}
\label{sec:background}
\vspace{-0.5em}

\textbf{Video foundation backbone.} We use a flow-matching video backbone with velocity field $v_\theta$. Given clean video latents $x_0$ and noise $\varepsilon \sim \mathcal{N}(0, I)$, the noisy latent at video noise level $\sigma_v \in [0, 1]$ is
$
x_{\sigma_v} = (1 - \sigma_v) x_0 + \sigma_v \varepsilon.
$
The video model predicts $v_\theta(x_{\sigma_v}, c, \sigma_v) \approx \varepsilon - x_0$ under text condition $c$:
\begin{align}
\mathcal{L}_{\mathrm{video}} = \mathbb{E}_{x_0, \varepsilon, \sigma_v}\bigl[\lVert v_\theta(x_{\sigma_v}, c, \sigma_v) - (\varepsilon - x_0) \rVert^2\bigr].
\label{eq:video-loss}
\end{align}
At target video noise level $\sigma_v$ during video denoising, we extract features from $k$-th block of the Cosmos-2.5 2B DiT backbone~\cite{ali2025world} (fixed at $k=20$ out of 28 total blocks following~\cite{pai2025mimic}):
$
F_\theta^{(k)}(x_{\sigma_v}, c, \sigma_v) \in \mathbb{R}^{T \times (H \times W) \times D},
$
where $t=0$ is the anchored clean observation (inpainted at every denoising step) and $t>0$ are predicted future latent frames.

\begin{figure}[t]
    \centering
    \includegraphics[width=1.0\linewidth]{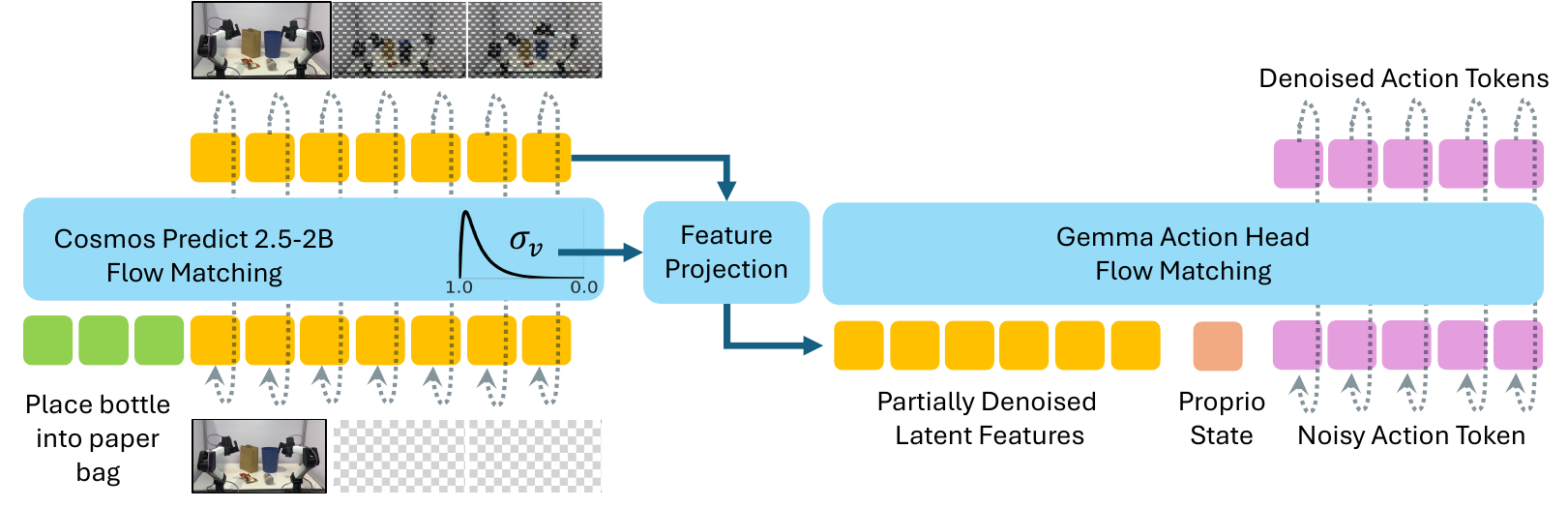}
    \caption{\textbf{Latent VAM architecture.} We replace the VLM backbone in $\pi_0$~\cite{black2024pi_0} with Cosmos-Predict 2.5 Video DiT~\cite{ali2025world}. From the current image and language instruction, plausible future rollout is partially denoised; intermediate DiT features and the video noise level are projected into the action head, then concatenated with proprioception and noisy action tokens for action denoising.}
    \label{fig:architecture}
    \vspace{-1.5em}
\end{figure}

\textbf{Action head integration.} The action head $\pi_\phi$ is a transformer policy (300M Gemma Transformer~\cite{Kamath2025Gemma3T, black2024pi_0, intelligence2025pi}) that consumes flattened video tokens, a projected proprioceptive token, and noisy action tokens as illustrated in~\autoref{fig:architecture}. For clean action sequence $a_0$, action noise $\varepsilon_a$, and action noise level $\sigma_a$, we define $a_{\sigma_a}=(1-\sigma_a)a_0+\sigma_a\varepsilon_a$ and train
\begin{align}
\mathcal{L}_{\mathrm{action}} = \mathbb{E}_{a_0, \varepsilon_a, \sigma_a}\bigl[\lVert v_\phi(a_{\sigma_a}, F_\theta^{(k)}, s, \sigma_a) - (\varepsilon_a - a_0) \rVert^2\bigr].
\end{align}
The action head uses bidirectional attention over the video prefix and causal attention over action tokens~\cite{black2024pi_0, intelligence2025pi}. We provide training-inference pseudocode in~\autoref{app:architecture_algorithm}.

\vspace{-0.5em}
\subsection{Design Axes}\label{sec:axes}
\vspace{-0.5em}

The design space is defined by two primary groups of axes: training-time configurations governing backbone adaptation, and inference-time parameters determining feature extraction behavior.

\textbf{Training axes.} (1) \emph{Finetuning strategy}: We compare parameter-efficient adaptation via Low-Rank Adaptation~\cite{hu2022lora} (training 2--6\% of parameters at $\eta = 10^{-4}$) against full finetuning of the video DiT backbone ($\eta = 10^{-6}$). (2) \emph{Training mode}: We evaluate \emph{joint training}, which optimizes the weighted objective $\mathcal{L}_{\mathrm{total}} = \mathcal{L}_{\mathrm{action}} + \lambda \mathcal{L}_{\mathrm{video}}$, enabling action gradients to directly reshape the DiT feature space. This is contrasted with \emph{two-stage training}, where the backbone is first finetuned on $\mathcal{L}_{\mathrm{video}}$ and subsequently frozen during action head optimization on $\mathcal{L}_{\mathrm{action}}$.

\textbf{Inference axes.} (1) \emph{Video noise level}: We extract features at varying values of $\sigma_v \in [0.0, 1.0]$ along the video denoising flow trajectory. Larger $\sigma_v$ yields noisier, less defined future latents, whereas smaller $\sigma_v$ produces highly denoised, semantically structured rollouts. (2) \emph{Prediction horizon}: We vary the temporal extent of the predicted rollout, $T$.

\begin{figure}[t]
    \centering
    \includegraphics[width=1\linewidth]{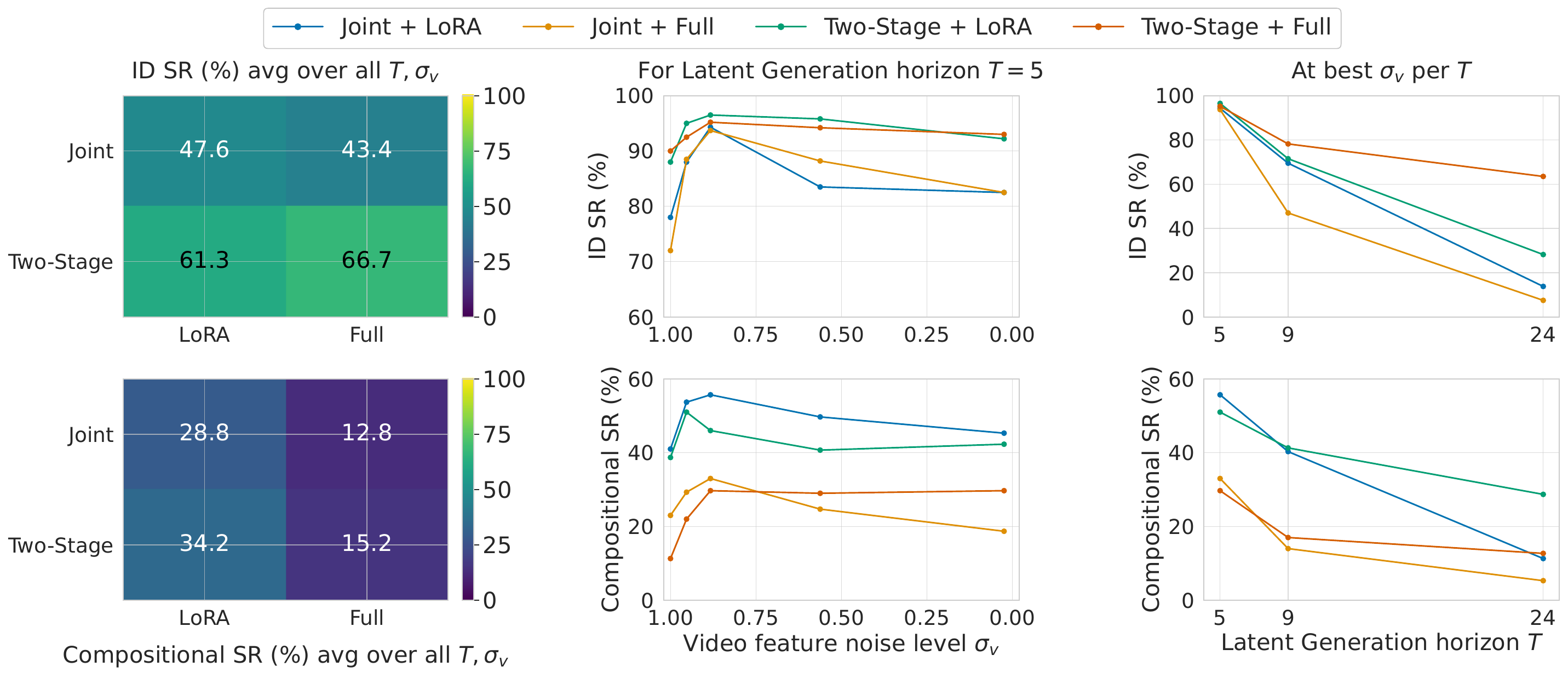}
    \caption{\textbf{Systematic exploration of the design space of VAMs.} (Left) We summarize the performance of finetuning strategy and coupled/decoupled video-action training. (Middle) We show the change of ID-OOD SR as the video noise level for feature extraction decreases. Performance peaks at an intermediate $\sigma_v$. (Right) We show the change of ID-OOD SR as the temporal horizon of videos increases. While longer temporal horizons better preserve object-goal relationships, this does not necessarily translate into higher SR. For each column, (Top) is ID SR and (Bottom) is OOD SR.}
    \label{fig:adaptation-results}
    \vspace{-1.5em}
\end{figure}

\vspace{-0.5em}
\section{Systematic Analysis of the Video-Action Generalization Gap}
\vspace{-0.5em}

We begin by asking: \emph{when does compositional generalization happen in VAMs?} To answer this, we conduct a wide range of ablation studies over the design axes introduced above: backbone adaptation, coupled/decoupled video-action training and video feature-extraction strategies.
These axes determine how the action head uses a language-conditioned, partially denoised latent rollout with a clean anchor frame ($t=0$). Hence, we use the design sweep to identify choices that preserve ID control while enabling compositional behavior.
Summarized in \autoref{fig:adaptation-results}, we discuss three key findings.

\textbf{Finetuning strategy changes ID and OOD success differently.} Full finetuning~\cite{ma2026dit4dit} can aggressively fit the in-domain tasks by specializing the latent rollout distribution to demonstrated trajectories. LoRA~\cite{pai2025mimic} restricts the update and tends to preserve more pretrained temporal structure. Crucially, ID and OOD performance do not scale proportionally: \textit{Joint-LoRA} yields the strongest OOD performance, while \textit{Two-Stage LoRA} is more robust when averaged across $\sigma_v$ and $T$.

\textbf{The best video noise level is neither fully noisy nor fully denoised.} At high noise levels ($\sigma_v \to 1.0$), future tokens lack coherent temporal structure, forcing the action head to anchor to the present state~\cite{pai2025mimic, ma2026dit4dit}. At very low noise levels ($\sigma_v \to 0.0$), future tokens become highly structured, but the policy is more vulnerable to confident hallucinated trajectories~\cite{li2025novaflow}. Performance therefore peaks at an intermediate noise level.

\begin{wrapfigure}{r}{0.35\textwidth}
\vspace{-2em}
  \begin{center}
    \includegraphics[trim={0 0.5cm 0 0},clip,width=\linewidth]{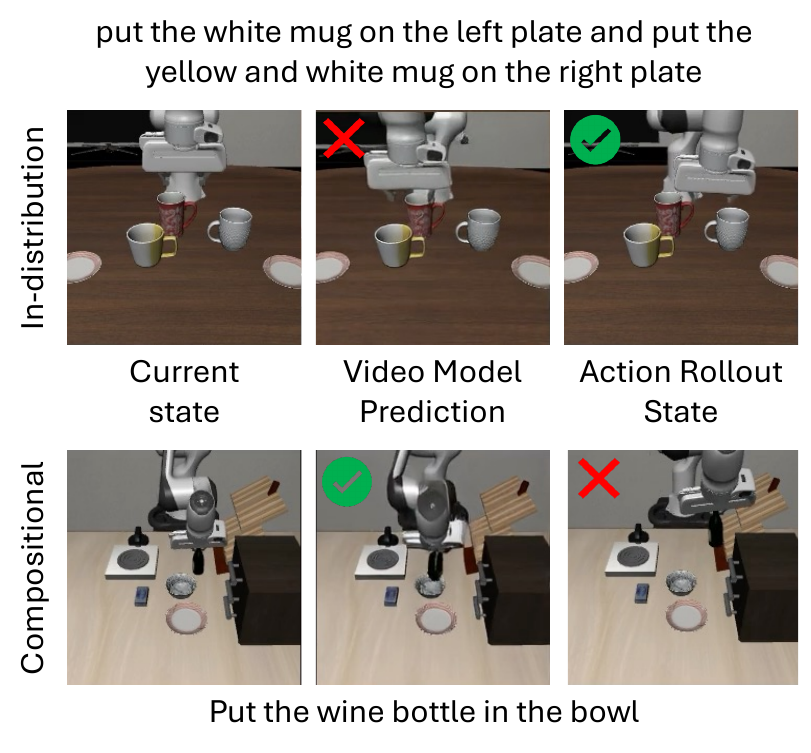}
  \end{center}
  \vspace{-0.5em}
  \caption{\textbf{Video-action disagreement.} (Top) The action head disagrees with an incorrect video prediction and executes correctly. (Bottom) The action head ignores a correct compositional prediction and executes incorrectly.}
  \label{fig:video_action_agreement}
  \vspace{-1.5em}
\end{wrapfigure}
\textbf{Longer video prediction improves video-level generalization, but does not necessarily improve action success.} Increasing the prediction horizon exposes more object-goal structure and can improve the video rollout's compositional plan qualitatively. However, longer horizons also increase the chance of implausible artifacts and do not automatically make the action head follow the correct plan. As a result, better video-level generalization does not always translate into higher task success. More details are provided in~\autoref{app:sim_design_analysis}.

\textbf{Motivating example: video-action disagreement.} The fragmented ablation landscape motivates a closer look at how the action head interprets latent video features. As shown in~\autoref{fig:video_action_agreement}, the same video rollout can be used or ignored depending on the task regime. In an ID LIBERO-long task (Top), the video model may hallucinate a wrong future while the action head still executes a correct memorized behavior. In a compositional OOD task (Bottom), the video model may propose the correct plan, but the action head ignores it and follows an ID trajectory. This failure mode cannot be diagnosed from the video rollout alone: a correct imagined future is only useful if the action policy follows it. Since video-feature utilization emerges implicitly from underlying design choices rather than explicit architectural constraints, we require a metric to quantify exactly how much the action head relies on the imagined future rollout relative to the current-state anchor.

\vspace{-0.5em}
\section{Understanding and Mitigating the VAG Gap via Temporal Ratio}
\vspace{-0.5em}


The design sweep and disagreement examples show that VAM success depends not only on whether the video backbone predicts a plausible future, but also on whether the action head routes through that future during action denoising. We therefore introduce the Temporal Ratio (TR), an attention-based measure of the action head's reliance on the predicted rollout. At each replan step, action tokens attend over video tokens containing the current frame and predicted future frames. Let $A_{\ell}$ be the head-averaged attention map from action tokens to video tokens at action-head layer $\ell$, after softmax. Partition video tokens into current-frame tokens $\mathcal{V}_0$ and future-frame tokens $\mathcal{V}_{+}$.

\vspace{-1em}
\begin{equation}
\label{eq:tr-definition}
\textit{Temporal Ratio:}\quad
    \mathrm{TR}_i^{(\ell)}
    =
    \frac{
        \sum_{q \in \mathcal{A}}
        \sum_{v \in \mathcal{V}_{+}}
        A_{\ell}(q,v)
    }{
        \sum_{q \in \mathcal{A}}
        \sum_{v \in \mathcal{V}_{0}}
        A_{\ell}(q,v)
    } = \frac{\text{Attention over future frames}}{\text{Attention over current frame}},
\end{equation}
\vspace{-1em}

where $\mathcal{A}$ denotes the action tokens. TR is high when actions rely on predicted futures and low when they anchor to the current observation. TR has a direct physical interpretation in VAMs, emerging as an artifact of the video denoising formulation: the first frame is injected as a clean conditioning frame at every video denoising step and serves as a high-fidelity observation anchor. Future frames, in contrast, are denoised from noisy latent targets, have lower signal-to-noise ratio~(SNR) than the anchor, and represent the model's imagined scene evolution. TR is useful because it is available from the policy's normal forward pass and can be tracked online across replan steps. It therefore provides a runtime view of when the policy enters a predictive mode, rather than a post-hoc score computed only after an episode terminates.

\begin{figure}[t]
    \centering
    \includegraphics[trim={0 0cm 0 0},clip,width=1\linewidth]{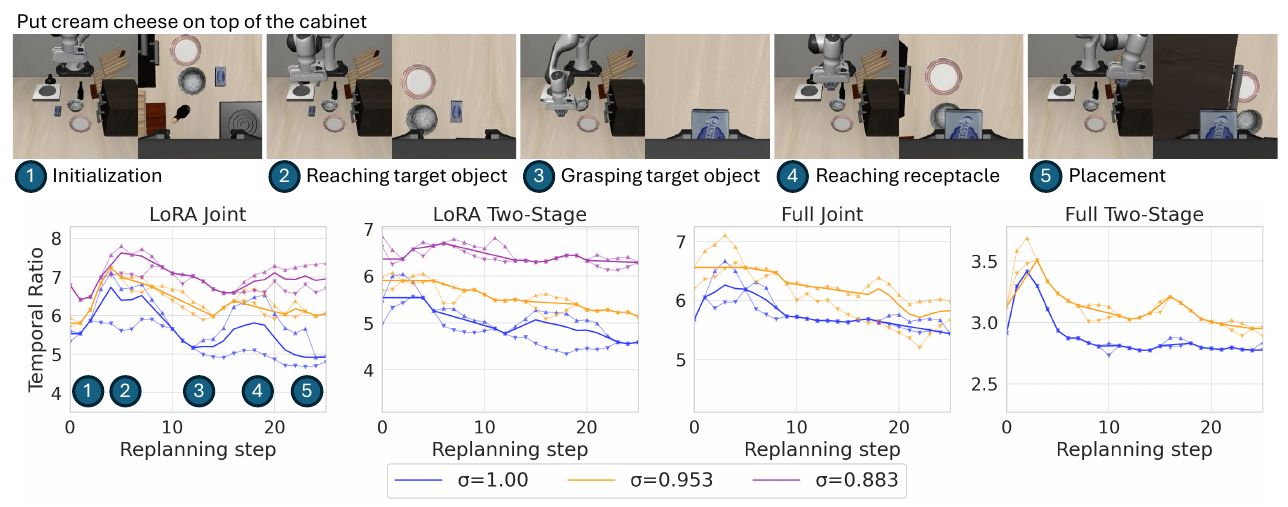}
    \caption{\textbf{Evolution of TR across episodes.} We illustrate the evolution of TR for each training category, for varying video noise levels and at $T=5$. Upper triangular markers denote TR values for successful episodes while lower triangular markers denote TR values for failed episodes. It is evident that: (1) successful episodes are consistently characterized by higher TR and (2) TR peaks during the planning phase and drops during precise grasping or placement, shown using a rollout example for the task ``put cream cheese on top of the cabinet.''}
    \label{fig:tr_evolution}
    \vspace{-1.5em}
\end{figure}

\vspace{-0.7em}
\subsection{TR Analysis in Simulation: Key Findings}
\label{sec:tr-analysis}
\vspace{-0.7em}

We measure TR on LIBERO across tasks, execution timesteps, design configurations, and success outcomes. This lets us diagnose whether the policy is using the clean anchor frame as a reactive controller or using future latents as a plan.

\textbf{TR increases with video denoising before feature extraction.} More video denoising improves the SNR and semantic structure of generated rollouts. As shown in~\autoref{fig:tr_evolution}, the action head reflects this change: high-noise level video features suppress TR because future tokens are too unstructured, while lower-noise level video features encourage attention to imagined futures.

\textbf{TR separates reactive and predictive phases of execution.} TR also changes within a rollout: (1) It rises during planning moments such as object or receptacle selection, where the policy must bind language, objects, and goals through the predicted future. (2) It drops during local manipulation, where grasping and placement require precise current-frame grounding. Thus, low TR is not always harmful; it is appropriate when the task phase demands reactive action generation.

This motivates using TR to detect the policy's execution phase at runtime and amplify compositional video conditioning primarily during predictive planning phases.
\vspace{-0.5em}
\subsection{Adaptive TR Guidance}
\label{sec:guidance}
\vspace{-0.5em}


Motivated by previous findings, we propose a new inference-time adaptive guidance method based on TR.
At each action denoising step, let $v_{\mathrm{cond}}$ be the standard conditional velocity. We compute two counterfactual directions that isolate complementary sources of compositional behavior.

\textbf{Language guidance:} We drop the text instruction ($c=\varnothing$) during video feature extraction and compute $v_{\mathrm{uncond}} = 
    v_\phi(a_{\sigma_a},F_{\varnothing}^{(k)},s,\sigma_a)$ where $F_{\varnothing}^{(k)} = F_\theta^{(k)}(x_{\sigma_v},\varnothing,\sigma_v)$. A guidance direction in action flow-velocity can be formulated as  $\Delta_{\mathrm{lang}}^v = v_{\mathrm{cond}} - v_{\mathrm{uncond}}$
    which isolates instruction-dependent action velocity and helps resolve ambiguous object-goal bindings.

\textbf{Plan guidance:} We query the video model with an extended horizon $T' > T$, truncate features back to the policy horizon $T$, and compute $v_{\mathrm{plan}}
    =
    v_\phi(a_{\sigma_a},F_{\mathrm{plan}}^{(k)},s,\sigma_a)$ where $F_{\mathrm{plan}}^{(k)}
    =
    F_\theta^{(k)}(x_{\sigma_v}^{(T')},c,\sigma_v)\big|_{\mathrm{first}\ T}$. A plan guidance direction can be formulated as $\Delta_{\mathrm{plan}}^v = v_{\mathrm{plan}} - v_{\mathrm{cond}}$
    which isolates long-horizon temporal structure while preserving the fixed action-head input horizon.

The combined guided update becomes
$v_{\mathrm{guided}} = v_{\mathrm{cond}} + w_{\mathrm{lang}, i} \Delta_{\mathrm{lang}}^v + w_{\mathrm{plan}, i} \Delta_{\mathrm{plan}}^v$.
Rather than using fixed weights throughout the episode, we scale the guidance strengths with the current TR. For $w \in \{w_{\mathrm{lang}}, w_{\mathrm{plan}}\}$, the weights can be dynamically updated as $w_{i} = w^{\mathrm{base}} s_i$ where 
$s_i~=~\mathrm{clip} \left( \frac{\mathrm{TR}_i}{\mathrm{TR}_0} - 1, 0, s_{\max} \right)$. Here, $s_i$ (1) is zero when the current TR drops below the initial value $\mathrm{TR}_0$, (2) grows as the policy shifts toward future-latent reliance, and (3) is capped by $s_{\max}$ to avoid over-guidance.
This concentrates intervention during planning phases and relaxes it during present-anchored manipulation, allowing guidance to improve compositional reasoning without hurting local precision while grasping or placement.
\vspace{-0.6em}
\section{Experimental Results}
\label{sec:results_sim}
\vspace{-0.7em}

The previous sections identify the failure mechanism behind the VAG gap: a VAM succeeds compositionally only when the video backbone predicts a useful future and the action head follows that future during planning. This section tests whether that diagnosis translates into measurable gains by evaluating: (1) the VAG gap in prior WAMs and VAMs, (2) whether our design choices reduce it, and (3) whether TR-Adaptive Guidance improves OOD success without hurting ID performance.

\textbf{Datasets.} We use the four standard LIBERO suites~\cite{Liu2023LIBEROBK}: \texttt{libero-spatial}, \texttt{libero-object}, \texttt{libero-goal}, and \texttt{libero-long} for simulation evaluation. Each suite contains 10 tasks, 50 demos per task. Following~\cite{kim2026cosmos}, the action head is trained on successful demos, while the video backbone finetuning uses the full unfiltered dataset. For real-world evaluation on bimanual YAM setup, we use 5600 episodes across 24 tasks from~\cite{abc2026}, including pick-and-place with dishes, snacks, bottles, and fruits; arranging objects on shelves; bimanual handover; t-shirt manipulation; and bottle-cap unscrewing. The multi-task dataset allows the policy to learn reusable object, receptacle, and manipulation factors. Observations use three 224$\times$168 RGB views (top, left wrist, right wrist) stacked horizontally, and the policy outputs a 14-dim bimanual action. More details are in~\autoref{app:real_world_setup}.

\begin{figure}[t]
    \centering
    \includegraphics[width=1\linewidth]{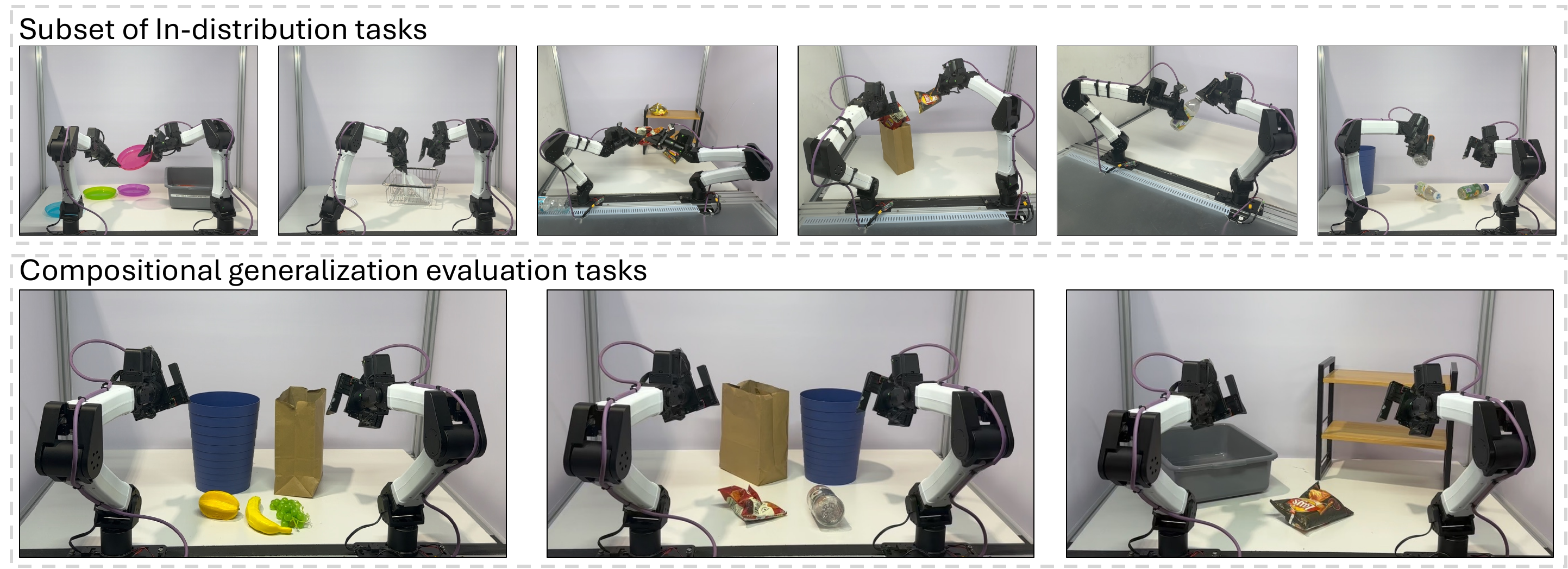}
    \caption{\textbf{Real-world setup.} (Top) We train on a bimanual YAM multi-task dataset with dish placement, snack sorting, bottle pick-and-place, and cap unscrewing. (Bottom) We show 3 test scenes with multiple target-object and receptacle candidates to evaluate compositional generalization.}
    \label{fig:real_world_evaluation}
    \vspace{-1.5em}
\end{figure}

\begin{table}[t]
\centering
\caption{LIBERO success rate (\%) across ID and OOD suites. Baseline ID numbers are from original papers where available; OOD results use released checkpoints with 50 episodes per task per suite and 3 seeds. ID success is saturated, while OOD success exposes the VAG gap.}
\label{tab:libero_complete}
\small
\setlength{\tabcolsep}{4pt}
\resizebox{\linewidth}{!}{%
\begin{tabular}{ll cccc c ccc c}
\toprule
& & \multicolumn{4}{c}{In-Distribution} & & \multicolumn{3}{c}{Out-of-Distribution} & \\
\cmidrule(lr){3-6} \cmidrule(lr){8-10}
Category & Method & Spatial & Object & Goal & Long & Avg ID & Spatial & Object & Goal & Avg OOD \\
\midrule
\multirow{2}{*}{\textit{VLA}}
    & $\pi_0$           & 96.8 & 98.8 & 95.8 & 85.2 & 94.2 & 0.7 & 0.3 & 4.3 & 1.7 \\
    & $\pi_{0.5}$         & 98.8 & 98.2 & 98.0 & 92.4 & 96.9 & 36.7 & 2.3 & 41.7 & 26.8 \\
\midrule
\multirow{6}{*}{\textit{WAM}/\textit{VAM}}
    & Cosmos-Policy     & 98.1 & 100.0 & 98.2 & 97.6 & \underline{98.5} & 30.7 & 0.3 & 0.7 & 10.5 \\
    & Fast-WAM          & 98.2 & 100.0 & 97.0 & 95.2 & 97.6 & 12.7 & 0.0 & 15.7 & 9.4 \\
    & Mimic-Video
    & 94.2 & 96.8 & 90.6 & -- & -- & -- & -- & -- & -- \\
    & DiT4DiT
    & 98.4 & 99.6 & 98.6 & 97.6 & \textbf{98.6} & 9.0 & 0.0 & 10.3 & 6.4 \\
\cmidrule{2-11}
& \multirow{1}{*}{\textit{Ours (best unguided)}}
    & 97.3 & 98.0 & 97.6 & 84.3 & 94.3 & 50.3 & 40.0 & 77.3 & \underline{55.7} \\
& \textit{Ours (best guided)}
    & 96.3 & 99.6 & 97.6 & 82.6 & 94.0 & 58.6 & 40.0 & 80.3 & \textbf{59.4} \\
\bottomrule
\end{tabular}
}
\vspace{-1em}
\end{table}

\textbf{Baselines in simulation and real-world study.} We compare against open-source VLAs~($\pi_0$~\cite{black2024pi_0}, $\pi_{0.5}$~\cite{intelligence2025pi}), WAMs~(Cosmos-Policy~\cite{kim2026cosmos}, Fast-WAM~\cite{yuan2026fast}), and VAMs~(Mimic-Video~\cite{pai2025mimic}, DiT4DiT~\cite{ma2026dit4dit}). All WAM/VAM baselines use the same Cosmos-Predict 2.5-2B video backbone, and the $\pi_0$-family VLA baselines are at a similar 3B VLM scale. This comparison separates generic VLA pretraining from explicit world-action modeling that modifies the video generative prior, and from latent video conditioning with a separate action head. For LIBERO, we evaluate released checkpoints on the OOD suites. For real-world tasks, we train $\pi_0$, $\pi_{0.5}$, and Cosmos-Policy on our bimanual YAM multi-task dataset with matched compute budgets: all equivalent to training for 1M steps on 32~NVIDIA\textregistered\ H200\texttrademark\ GPUs. More details are in~\autoref{app:real_world_setup}.

\vspace{-0.5em}
\subsection{Compositional generalization evaluation testbed}
\vspace{-0.5em}

\textbf{LIBERO.} We use the compositional OOD setup from~\citet{li2025task}. These tasks recombine familiar objects, receptacles, layouts, and goals into unseen task sequences. The object and destination may each appear in training, but the connecting trajectory is novel, so success requires stitching sub-trajectories learned from disjoint ID tasks. \texttt{libero-goal-ood} and \texttt{libero-spatial-ood} test novel object-destination or layout-destination combinations while \texttt{libero-object-ood} requires semantic understanding of the object categories. More details are in~\autoref{app:sim_task_setup}.

\textbf{Bimanual YAM.} We construct three OOD scenes with multiple target and receptacle candidates as shown in~\autoref{fig:real_world_evaluation}. Tasks 1--2 test unseen fruit-receptacle bindings and require selecting the correct receptacle. Tasks 3--6 require choosing between snacks/bottles and paper bag/bin, mixing seen and unseen pairings under a multi-receptacle shift. Tasks 7--8 test whether the policy overfits to the seen snack-shelf behavior when bin and shelf are both present. More details are in~\autoref{app:real_world_setup}.


\vspace{-0.5em}
\subsection{Key Findings}
\vspace{-0.5em}

\textbf{The VAG gap appears across baselines.} \autoref{tab:libero_complete} compares ID and OOD success\footnote{Mimic-Video~\cite{pai2025mimic} released checkpoints per ID task suite, making OOD comparison unfair.}. ID performance is saturated for most methods ($>95\%$), but compositional OOD success collapses for prior WAMs/VAMs. Among baselines, $\pi_{0.5}$ has the strongest OOD average, suggesting that strong VLA pretraining can reduce but not eliminate the gap. Our best unguided VAM extracts features at $\sigma_v=0.883$ with horizon $T=5$ (17 frames at 10 fps), while the action head predicts 16 actions at 10 Hz. This adds marginal compute while improving avg. OOD to 55.7\%. Adaptive guidance further raises avg. OOD to 59.4\%, more than $5\times$ prior VAM performance.

\textbf{TR-based adaptive guidance mitigates the gap when video futures are useful.} Guidance preserves most ID performance but slightly hurts LIBERO-long, indicating that stronger video conditioning can be harmful when predicted futures are unreliable. On compositional OOD suites, however, guidance improves avg. OOD by $\approx 6\%$. This supports the central mechanism: action-level generalization requires both a useful video rollout and sufficient action-head reliance on that rollout. If video features generalize but TR is low, the action head ignores the compositional plan; if TR is high but the video future is wrong, guidance can amplify the wrong behavior.

\begin{table}[t]
  \centering
  \caption{We compare the performance of baselines and our VAM with optimal design choices, with and without guidance, across 3 compositional evaluation scenes and 8 tasks. Tasks 1--2 are evaluated with 5 trials with 3 fruits~(total 15 trials), and Tasks 3--8 are evaluated with 5 trials each. The tasks are designed such that a policy capable of precise manipulation but no generalization will have an avg. performance of 50\%. Performance $>50\%$ illustrates compositional generalization capabilities.}
  \label{tab:yam_complete}
  \small
  \setlength{\tabcolsep}{6.5pt}
  \begin{tabular}{l |cc|cccc|cc| c}
  \toprule
  \multirow{2}{*}{Method} & \multicolumn{2}{c|}{Scene 1} & \multicolumn{4}{c|}{Scene 2} & \multicolumn{2}{c|}{Scene 3} & \multirow{2}{*}{Avg} \\
  \cmidrule{2-9}
   & Task 1 & Task 2 & Task 3 & Task 4 & Task 5 & Task 6 & Task 7 & Task 8 &  \\
  \midrule
  $\pi_{0}$           & 66.7 & 6.7  & 20.0  & 20.0 & 0.0  & 40.0  & 40.0  & 100.0 & 36.7 \\
  $\pi_{0.5}$         & 80.0 & 20.0 & 100.0 & 40.0 & 20.0 & 20.0  & 100.0 & 80.0  & 55.0 \\
  Cosmos-Policy       & 8.3  & 8.3  & 15.0  & 10.0 & 10.0 & 15.0  & 25.0  & 25.0  & 12.5 \\
  \textit{Ours (unguided)}  & 80.0 & 53.3 & 100.0 & 60.0 & 40.0 & 100.0 & 100.0 & 60.0  & \underline{71.7} \\
  \textit{Ours (guided)}    & 80.0 & 73.3 & 100.0 & 80.0 & 80.0 & 100.0 & 100.0 & 80.0  & \textbf{83.3} \\
  \bottomrule
  \end{tabular}
  \vspace{-1em}
\end{table}

\textbf{TR-based adaptive guidance boosts real-world performance on bimanual YAM.} We transfer the best LIBERO design, LoRA backbone adaptation with joint training, to YAM. Video features are extracted at $\sigma_v=0.883$ with horizon $T=5$ (17 frames at 10 fps), while the action head predicts 50 actions at 30 Hz. \autoref{tab:yam_complete} shows that our VAM outperforms baselines\footnote{Cosmos-Policy did not train well with the open-source codebase on our multi-task dataset despite sharing the same video backbone, suggesting that large multi-task WAM training remains an open problem.}, and adaptive guidance further improves average success from 71.7\% to 83.3\%. Extracting features at the first denoising step ($\sigma_v=1.0$), as in prior VAMs~\cite{pai2025mimic, ma2026dit4dit}, fails on all YAM tasks, highlighting the importance of structured, partially denoised video latents in multi-task VAMs.

\section{Related Work}
\label{sec:related_work}

\textbf{Generative video foundation models as backbone for policy learning.}
Generative video models~\citep{ali2025world, wan2025wan, VeoTeam2025} provide spatio-temporal priors absent from standard VLAs~\cite{kim2024openvla, black2024pi_0, intelligence2025pi, intelligence2026pi, bjorck2025gr00t, team2025gemini, fang2026molmoact2}. Robotics work transfers these priors through imagine-then-execute methods~\citep{jang2025dreamgen, li2025novaflow, du2023learning, du2024video, ko2024learning}, WAMs that jointly model video and action~\citep{kim2026cosmos, ye2026world, chen2025large, li2026causal, deng2025emerging, wu2024unleashing, li2025unified, zhu2025unified, zheng2025flare, zhao2025cot, tharwat2025latent, hu2024video, liang2025video}, and latent VAMs that feed video features to an action head~\citep{gpc, luo2025grounding, pai2025mimic, ma2026dit4dit}. Complementing Fast-WAM~\citep{yuan2026fast} and HarmoWAM~\citep{feng2026harmowam}, we study whether the action head actually uses future-predictive video features or retreats to present-state anchoring.

\textbf{Studying compositional generalization for foundation-model-based policies.}
Compositional generalization~\cite{Okawa2023CompositionalAE, wiedemer2023compositional, farid2025drives, chen2025robohiman} is necessary because combinatorial demonstrations do not scale. Prior VLA studies identify spatial overfitting~\citep{li2025task}, modality imbalance~\citep{fang2026vision}, and brittle linguistic grounding~\citep{fei2025libero, zhou2025libero}; larger frontier systems reduce but do not remove the gap~\citep{intelligence2026pi, ye2026world}. We focus on video-conditioned policies, where the central question is whether generative temporal priors survive action finetuning and improve novel sequence compositions.
\section{Conclusion}
\label{sec:conclusion}

We study the VAG gap: when VAMs fail to inherit the compositional priors of their generative video backbones. Temporal Ratio (TR) reveals that successful compositional behavior appears when the action head shifts toward future-predictive latents during planning, then returns to present anchoring for local manipulation. TR also enables TR-Adaptive Guidance, an inference-time intervention that amplifies compositional video conditioning during planning phases. Across LIBERO and real-robot tasks, this framework clarifies VAM behavior and mitigates the OOD generalization gap. We will open-source our study codebase and trained models after double-blind review.

\vspace{-0.5em}
\paragraph{Limitations.} Several limitations remain: (1) the presented study remains confined to a specific class of VAMs and is bottlenecked by the physical plausibility and compositional reach of the underlying video model, especially under distribution shifts where generated futures may become confident but infeasible. In the latter case, adaptive guidance will amplify incorrect guidance signals. (2) extra denoising steps and guidance add inference cost, reducing the inference frequency. More details are in~\autoref{app:real_world_setup}.


\clearpage


\bibliography{example}  

@article{Okawa2023CompositionalAE,
  title={Compositional Abilities Emerge Multiplicatively: Exploring Diffusion Models on a Synthetic Task},
  author={Maya Okawa and Ekdeep Singh Lubana and Robert P. Dick and Hidenori Tanaka},
  journal={ArXiv},
  year={2023},
  volume={abs/2310.09336},
  url={https://api.semanticscholar.org/CorpusID:264146105}
}

@article{wiedemer2023compositional,
  title={Compositional generalization from first principles},
  author={Wiedemer, Thadd{\"a}us and Mayilvahanan, Prasanna and Bethge, Matthias and Brendel, Wieland},
  journal={Advances in Neural Information Processing Systems},
  volume={36},
  pages={6941--6960},
  year={2023}
}

@article{farid2025drives,
  title={What Drives Compositional Generalization in Visual Generative Models?},
  author={Farid, Karim and Sahay, Rajat and Alnaggar, Yumna Ali and Schrodi, Simon and Fischer, Volker and Schmid, Cordelia and Brox, Thomas},
  journal={arXiv preprint arXiv:2510.03075},
  year={2025}
}

@article{li2025task,
  title={Task Reconstruction and Extrapolation for $\pi_0$ using Text Latent},
  author={Li, Quanyi},
  journal={arXiv preprint arXiv:2505.03500},
  year={2025}
}

@article{fang2026vision,
  title={When vision overrides language: Evaluating and mitigating counterfactual failures in VLAs},
  author={Fang, Yu and Feng, Yuchun and Jing, Dong and Liu, Jiaqi and Yang, Yue and Wei, Zhenyu and Szafir, Daniel and Ding, Mingyu},
  journal={arXiv preprint arXiv:2602.17659},
  year={2026}
}

@article{fei2025libero,
  title={Libero-plus: In-depth robustness analysis of vision-language-action models},
  author={Fei, Senyu and Wang, Siyin and Shi, Junhao and Dai, Zihao and Cai, Jikun and Qian, Pengfang and Ji, Li and He, Xinzhe and Zhang, Shiduo and Fei, Zhaoye and others},
  journal={arXiv preprint arXiv:2510.13626},
  year={2025}
}

@article{zhou2025libero,
  title={LIBERO-PRO: Towards Robust and Fair Evaluation of Vision-Language-Action Models Beyond Memorization},
  author={Zhou, Xueyang and Xu, Yangming and Tie, Guiyao and Chen, Yongchao and Zhang, Guowen and Chu, Duanfeng and Zhou, Pan and Sun, Lichao},
  journal={arXiv preprint arXiv:2510.03827},
  year={2025}
}

@article{kim2024openvla,
  title={Openvla: An open-source vision-language-action model},
  author={Kim, Moo Jin and Pertsch, Karl and Karamcheti, Siddharth and Xiao, Ted and Balakrishna, Ashwin and Nair, Suraj and Rafailov, Rafael and Foster, Ethan and Lam, Grace and Sanketi, Pannag and others},
  journal={arXiv preprint arXiv:2406.09246},
  year={2024}
}

@article{black2024pi_0,
  title={$\pi_0$: A Vision-Language-Action Flow Model for General Robot Control},
  author={Black, Kevin and Brown, Noah and Driess, Danny and Esmail, Adnan and Equi, Michael and Finn, Chelsea and Fusai, Niccolo and Groom, Lachy and Hausman, Karol and Ichter, Brian and others},
  journal={arXiv preprint arXiv:2410.24164},
  year={2024}
}

@article{intelligence2025pi,
  title={$\pi_{0.5}$: a Vision-Language-Action Model with Open-World Generalization},
  author={Intelligence, Physical and Black, Kevin and Brown, Noah and Darpinian, James and Dhabalia, Karan and Driess, Danny and Esmail, Adnan and Equi, Michael and Finn, Chelsea and Fusai, Niccolo and others},
  journal={arXiv preprint arXiv:2504.16054},
  year={2025}
}

@article{intelligence2026pi,
  title={$\pi_{0.7}$: a Steerable Generalist Robotic Foundation Model with Emergent Capabilities},
  author={Intelligence, Physical and Ai, Bo and Amin, Ali and Aniceto, Raichelle and Balakrishna, Ashwin and Balke, Greg and Black, Kevin and Bokinsky, George and Cao, Shihao and Charbonnier, Thomas and others},
  journal={arXiv preprint arXiv:2604.15483},
  year={2026}
}

@article{bjorck2025gr00t,
  title={Gr00t n1: An open foundation model for generalist humanoid robots},
  author={Bjorck, Johan and Casta{\~n}eda, Fernando and Cherniadev, Nikita and Da, Xingye and Ding, Runyu and Fan, Linxi and Fang, Yu and Fox, Dieter and Hu, Fengyuan and Huang, Spencer and others},
  journal={arXiv preprint arXiv:2503.14734},
  year={2025}
}

@article{team2025gemini,
  title={Gemini robotics: Bringing ai into the physical world},
  author={Team, Gemini Robotics and Abeyruwan, Saminda and Ainslie, Joshua and Alayrac, Jean-Baptiste and Arenas, Montserrat Gonzalez and Armstrong, Travis and Balakrishna, Ashwin and Baruch, Robert and Bauza, Maria and Blokzijl, Michiel and others},
  journal={arXiv preprint arXiv:2503.20020},
  year={2025}
}

@article{fang2026molmoact2,
  title={MolmoAct2: Action Reasoning Models for Real-world Deployment},
  author={Fang, Haoquan and Duan, Jiafei and Clay, Donovan and Wang, Sam and Liu, Shuo and Huang, Weikai and Fan, Xiang and Tsai, Wei-Chuan and Chen, Shirui and Wang, Yi Ru and others},
  journal={arXiv preprint arXiv:2605.02881},
  year={2026}
}

@article{ye2026world,
  title={World action models are zero-shot policies},
  author={Ye, Seonghyeon and Ge, Yunhao and Zheng, Kaiyuan and Gao, Shenyuan and Yu, Sihyun and Kurian, George and Indupuru, Suneel and Tan, You Liang and Zhu, Chuning and Xiang, Jiannan and others},
  journal={arXiv preprint arXiv:2602.15922},
  year={2026}
}

@article{chen2025large,
  title={Large video planner enables generalizable robot control},
  author={Chen, Boyuan and Zhang, Tianyuan and Geng, Haoran and Zhang, Caiyi and Li, Peihao and Song, Kiwhan and Freeman, William T and Malik, Jitendra and Abbeel, Pieter and Tedrake, Russ and others},
  journal={arXiv preprint arXiv:2512.15840},
  year={2025}
}

@article{kim2026cosmos,
  title={Cosmos policy: Fine-tuning video models for visuomotor control and planning},
  author={Kim, Moo Jin and Gao, Yihuai and Lin, Tsung-Yi and Lin, Yen-Chen and Ge, Yunhao and Lam, Grace and Liang, Percy and Song, Shuran and Liu, Ming-Yu and Finn, Chelsea and others},
  journal={arXiv preprint arXiv:2601.16163},
  year={2026}
}

@article{li2026causal,
  title={Causal World Modeling for Robot Control},
  author={Li, Lin and Zhang, Qihang and Luo, Yiming and Yang, Shuai and Wang, Ruilin and Han, Fei and Yu, Mingrui and Gao, Zelin and Xue, Nan and Zhu, Xing and others},
  journal={arXiv preprint arXiv:2601.21998},
  year={2026}
}

@article{deng2025emerging,
  title={Emerging properties in unified multimodal pretraining},
  author={Deng, Chaorui and Zhu, Deyao and Li, Kunchang and Gou, Chenhui and Li, Feng and Wang, Zeyu and Zhong, Shu and Yu, Weihao and Nie, Xiaonan and Song, Ziang and others},
  journal={arXiv preprint arXiv:2505.14683},
  year={2025}
}

@inproceedings{wu2024unleashing,
  title={Unleashing large-scale video generative pre-training for visual robot manipulation},
  author={Wu, Hongtao and Jing, Ya and Cheang, Chilam and Chen, Guangzeng and Xu, Jiafeng and Li, Xinghang and Liu, Minghuan and Li, Hang and Kong, Tao},
  booktitle={International Conference on Learning Representations},
  volume={2024},
  pages={10641--10662},
  year={2024}
}

@article{li2025unified,
  title={Unified video action model},
  author={Li, Shuang and Gao, Yihuai and Sadigh, Dorsa and Song, Shuran},
  journal={arXiv preprint arXiv:2503.00200},
  year={2025}
}

@article{zhu2025unified,
  title={Unified world models: Coupling video and action diffusion for pretraining on large robotic datasets},
  author={Zhu, Chuning and Yu, Raymond and Feng, Siyuan and Burchfiel, Benjamin and Shah, Paarth and Gupta, Abhishek},
  journal={arXiv preprint arXiv:2504.02792},
  year={2025}
}

@article{zheng2025flare,
  title={Flare: Robot learning with implicit world modeling},
  author={Zheng, Ruijie and Wang, Jing and Reed, Scott and Bjorck, Johan and Fang, Yu and Hu, Fengyuan and Jang, Joel and Kundalia, Kaushil and Lin, Zongyu and Magne, Loic and others},
  journal={arXiv preprint arXiv:2505.15659},
  year={2025}
}

@inproceedings{zhao2025cot,
  title={Cot-vla: Visual chain-of-thought reasoning for vision-language-action models},
  author={Zhao, Qingqing and Lu, Yao and Kim, Moo Jin and Fu, Zipeng and Zhang, Zhuoyang and Wu, Yecheng and Li, Zhaoshuo and Ma, Qianli and Han, Song and Finn, Chelsea and others},
  booktitle={Proceedings of the Computer Vision and Pattern Recognition Conference},
  pages={1702--1713},
  year={2025}
}

@article{tharwat2025latent,
  title={Latent action pretraining through world modeling},
  author={Tharwat, Bahey and Nasser, Yara and Abouzeid, Ali and Reid, Ian},
  journal={arXiv preprint arXiv:2509.18428},
  year={2025}
}

@article{liang2025video,
  title={Video generators are robot policies},
  author={Liang, Junbang and Tokmakov, Pavel and Liu, Ruoshi and Sudhakar, Sruthi and Shah, Paarth and Ambrus, Rares and Vondrick, Carl},
  journal={arXiv preprint arXiv:2508.00795},
  year={2025}
}

@article{hu2024video,
  title={Video prediction policy: A generalist robot policy with predictive visual representations},
  author={Hu, Yucheng and Guo, Yanjiang and Wang, Pengchao and Chen, Xiaoyu and Wang, Yen-Jen and Zhang, Jianke and Sreenath, Koushil and Lu, Chaochao and Chen, Jianyu},
  journal={arXiv preprint arXiv:2412.14803},
  year={2024}
}

@ARTICLE{gpc,
  author={Qi, Han and Yin, Haocheng and Zhu, Aris and Du, Yilun and Yang, Heng},
  journal={IEEE Robotics and Automation Letters}, 
  title={Inference-Time Enhancement of Generative Robot Policies via Predictive World Modeling}, 
  year={2026},
  volume={11},
  number={5},
  pages={5534-5541},
  doi={10.1109/LRA.2026.3673995}}

@inproceedings{luo2025grounding,
  title={Grounding video models to actions through goal conditioned exploration},
  author={Luo, Yunhao and Du, Yilun},
  booktitle={International Conference on Learning Representations},
  volume={2025},
  pages={92200--92232},
  year={2025}
}

@article{pai2025mimic,
  title={mimic-video: Video-action models for generalizable robot control beyond vlas},
  author={Pai, Jonas and Achenbach, Liam and Montesinos, Victoriano and Forrai, Benedek and Mees, Oier and Nava, Elvis},
  journal={arXiv preprint arXiv:2512.15692},
  year={2025}
}

@article{ma2026dit4dit,
  title={Dit4dit: Jointly modeling video dynamics and actions for generalizable robot control},
  author={Ma, Teli and Zheng, Jia and Wang, Zifan and Jiang, Chunli and Cui, Andy and Liang, Junwei and Yang, Shuo},
  journal={arXiv preprint arXiv:2603.10448},
  year={2026}
}

@article{yuan2026fast,
  title={Fast-WAM: Do World Action Models Need Test-time Future Imagination?},
  author={Yuan, Tianyuan and Dong, Zibin and Liu, Yicheng and Zhao, Hang},
  journal={arXiv preprint arXiv:2603.16666},
  year={2026}
}

@article{feng2026harmowam,
  title={HarmoWAM: Harmonizing Generalizable and Precise Manipulation via Adaptive World Action Models},
  author={Feng, Qiuxuan and Yu, Jiale and Liu, Jiaming and Jia, Yueru and Wu, Zhuangzhe and Chen, Hao and Qian, Zezhong and Gu, Shuo and Jia, Peng and Ma, Siwei and others},
  journal={arXiv preprint arXiv:2605.10942},
  year={2026}
}

@article{jang2025dreamgen,
  title={Dreamgen: Unlocking generalization in robot learning through video world models},
  author={Jang, Joel and Ye, Seonghyeon and Lin, Zongyu and Xiang, Jiannan and Bjorck, Johan and Fang, Yu and Hu, Fengyuan and Huang, Spencer and Kundalia, Kaushil and Lin, Yen-Chen and others},
  journal={arXiv preprint arXiv:2505.12705},
  year={2025}
}

@article{li2025novaflow,
  title={Novaflow: Zero-shot manipulation via actionable flow from generated videos},
  author={Li, Hongyu and Sun, Lingfeng and Hu, Yafei and Ta, Duy and Barry, Jennifer and Konidaris, George and Fu, Jiahui},
  journal={arXiv preprint arXiv:2510.08568},
  year={2025}
}

@article{du2023learning,
  title={Learning universal policies via text-guided video generation},
  author={Du, Yilun and Yang, Sherry and Dai, Bo and Dai, Hanjun and Nachum, Ofir and Tenenbaum, Josh and Schuurmans, Dale and Abbeel, Pieter},
  journal={Advances in neural information processing systems},
  volume={36},
  pages={9156--9172},
  year={2023}
}

@inproceedings{du2024video,
  title={Video language planning},
  author={Du, Yilun and Yang, Sherry and Florence, Pete and Xia, Fei and Wahid, Ayzaan and Sermanet, Pierre and Yu, Tianhe and Abbeel, Pieter and Tenenbaum, Joshua B and Kaelbling, Leslie and others},
  booktitle={International Conference on Learning Representations},
  volume={2024},
  pages={31138--31155},
  year={2024}
}

@inproceedings{ko2024learning,
  title={Learning to act from actionless videos through dense correspondences},
  author={Ko, Po-Chen and Mao, Jiayuan and Du, Yilun and Sun, Shao-Hua and Tenenbaum, Joshua B},
  booktitle={International Conference on Learning Representations},
  volume={2024},
  pages={40938--40958},
  year={2024}
}

@techreport{VeoTeam2025,
  title       = {Veo: a text-to-video generation system},
  author      = {{Google DeepMind Veo Team}},
  year        = {2025},
  institution = {Google DeepMind},
  url         = {https://storage.googleapis.com/deepmind-media/veo/Veo-3-Tech-Report.pdf}
}

@article{wan2025wan,
  title={Wan: Open and advanced large-scale video generative models},
  author={Wan, Team and Wang, Ang and Ai, Baole and Wen, Bin and Mao, Chaojie and Xie, Chen-Wei and Chen, Di and Yu, Feiwu and Zhao, Haiming and Yang, Jianxiao and others},
  journal={arXiv preprint arXiv:2503.20314},
  year={2025}
}

@article{ali2025world,
  title={World simulation with video foundation models for physical ai},
  author={Ali, Arslan and Bai, Junjie and Bala, Maciej and Balaji, Yogesh and Blakeman, Aaron and Cai, Tiffany and Cao, Jiaxin and Cao, Tianshi and Cha, Elizabeth and Chao, Yu-Wei and others},
  journal={arXiv preprint arXiv:2511.00062},
  year={2025}
}

@article{Liu2023LIBEROBK,
  title={LIBERO: Benchmarking Knowledge Transfer for Lifelong Robot Learning},
  author={Bo Liu and Yifeng Zhu and Chongkai Gao and Yihao Feng and Qian Liu and Yuke Zhu and Peter Stone},
  journal={ArXiv},
  year={2023},
  volume={abs/2306.03310},
  url={https://api.semanticscholar.org/CorpusID:259089508}
}

@article{Kamath2025Gemma3T,
  title={Gemma 3 Technical Report},
  author={Gemma Team},
  journal={ArXiv},
  year={2025},
  volume={abs/2503.19786},
  url={https://api.semanticscholar.org/CorpusID:277313563}
}

@article{hu2022lora,
  title={Lora: Low-rank adaptation of large language models.},
  author={Hu, Edward J and Shen, Yelong and Wallis, Phillip and Allen-Zhu, Zeyuan and Li, Yuanzhi and Wang, Shean and Wang, Liang and Chen, Weizhu and others},
  journal={Iclr},
  volume={1},
  number={2},
  pages={3},
  year={2022}
}

@article{lipman2022flow,
  title={Flow matching for generative modeling},
  author={Lipman, Yaron and Chen, Ricky TQ and Ben-Hamu, Heli and Nickel, Maximilian and Le, Matt},
  journal={arXiv preprint arXiv:2210.02747},
  year={2022}
}

@article{chen2025robohiman,
  title={RoboHiMan: A Hierarchical Evaluation Paradigm for Compositional Generalization in Long-Horizon Manipulation},
  author={Chen, Yangtao and Chen, Zixuan and Chan, Nga Teng and Chen, Junting and Yin, Junhui and Shi, Jieqi and Gao, Yang and Li, Yong-Lu and Huo, Jing},
  journal={arXiv preprint arXiv:2510.13149},
  year={2025}
}

@misc{i2rt_yam_arm,
  title={{YAM} 6-{DOF} Arm},
  author={{I2RT}},
  year={2025},
  howpublished={\url{https://i2rt.com/products/yam-6-dof-arm}},
  note={Accessed 2026-06-03}
}

@article{abc2026,
  title   = {Scalable Behavior Cloning with Open Data, Training, and Evaluation},
  author  = {Allshire, Arthur and Singh, Himanshu Gaurav and Singh, Ritvik and Rashid, Adam and Choi, Hongsuk and McAllister, David and Yu, Justin and Chen, Yiyuan and Huang, Huang and Abbeel, Pieter and Chen, Xi and Duan, Rocky and Isola, Phillip and Malik, Jitendra and Shentu, Fred and Shi, Guanya and Wu, Philipp and Kanazawa, Angjoo},
  year    = {2026},
  journal = {arXiv preprint},
  url     = {https://abc.bot/},
}

\newpage

\clearpage
\appendix
\phantomsection
{
\begingroup
\hypersetup{linkcolor=black}
\tableofcontents
\endgroup
}
\newpage

All videos are included in supplementary and \url{https://umishra.me/temporal-ratio/}.

\section{Architecture and Algorithm Details}
\label{app:architecture_algorithm}

\subsection{Latent VAM Architecture}
\label{app:latent-vam-architecture}

\paragraph{Video backbone.} We instantiate $F_\theta$ as the 28-block Cosmos-Predict2.5-2B DiT~\cite{ali2025world} (${\sim}2.1$B parameters). Inputs are VAE-encoded video latents from the Cosmos-Tokenizer ($4{\times}$ temporal and $8{\times}$ spatial compression), patchified by a Rearrange + Linear embedder with patch size $1{\times}2{\times}2$, and position-encoded with 3D rotary embeddings. Each block contains self-attention over video tokens, cross-attention to text, a linear module ($2048{\to}8192{\to}2048$), and Adaptive LayerNorm (shift/scale/gate) conditioned on the diffusion timestep. Self-attention is 2048-dim with 16 heads; cross-attention queries from video ($2048$) and projects keys/values from text ($1024{\to}2048$). Text is encoded by Cosmos-Reason1-7B (Qwen2.5-VL-7B), giving task prompt embeddings of shape $(B, 512, 100352)$ that are projected once through a shared linear layer $(100352{\to}1024)$ and reused as the same cross-attention embedding in all 28 blocks (no per-block text modulation).

\paragraph{Action head.} The action head $\pi_\phi$ is the 18-layer Gemma expert from $\pi_0$~\cite{black2024pi_0, intelligence2025pi, Kamath2025Gemma3T} (${\sim}300$M parameters, hidden dim $1024$). Attention is multi-query: 8 query heads and 1 shared key/value head with head dimension $256$. Each transformer layer uses a GeGLU MLP  and Adaptive RMSNorm whose dense projection $\mathbb{R}^{1024}{\to}\mathbb{R}^{3072}$ produces a shift/scale/gate triple conditioned on the action diffusion timestep $\sigma_a$, with the gate modulating the residual contribution of each sublayer. The token sequence concatenates flattened video tokens (per latent frame), a single state token, and action horizon $A_H$ ($=17$ for sim and $=50$ fo real) noisy action tokens; attention follows a prefix-LM mask~\cite{black2024pi_0}: bidirectional within the video prefix, causal at the state and first-action boundaries, and bidirectional among the remaining action tokens.

\paragraph{Video-feature interface.} At inference, we extract $F_\theta^{(k)}(x_{\sigma_v}, c, \sigma_v) \in \mathbb{R}^{T \times (H\times W) \times D}$ with $k{=}20$, $D{=}2048$, and $(H, W){=}(14, 28)$ from the $448{\times}224$ observation (primary + wrist camera, concatenated horizontally). Each spatial $1{\times}2{\times}2$ neighborhood (4 tokens, $4{\times}2048{=}8192$ features) is projected to the action-head dimension ($1024$) by a linear layer, yielding $98$ video tokens per latent frame. The video noise level $\sigma_v$ is encoded with a Fourier feature MLP and applied once through a video-only Adaptive layer norm (shift+scale, no gate) before the tokens enter the Gemma transformer, so that each video token carries an explicit ``noise level'' tag. Proprioception ($8$-dim) is padded to $32$ following $\pi-$models~\cite{black2024pi_0} and then embedded to $1024$ via \texttt{state\_proj}; noisy action chunks ($A_H$ steps $\times$ $A_D$ dims, internally padded to dimension $32$, where $A_D=7$ for sim and $=14$ for real) are likewise embedded as well, and then finally decoded into original dimensions to get the flow velocity output. The language instruction is consumed only by the video backbone via cross-attention; the action head sees video features, state, and the action timestep alone.

\subsection{Training and Inference Procedure}
\label{app:training-inference-procedure}

\paragraph{Video denoising.} The video backbone is trained with the rectified-flow loss of ~\autoref{eq:video-loss} under $\sigma_v \sim \mathcal{T}_v = \mathcal{U}[0, 1]$, predicting velocity $\varepsilon_v - x_0$ from $x_{\sigma_v} = (1{-}\sigma_v) x_0 + \sigma_v \varepsilon_v$. At inference we discretize the flow integral $x_0 = x_1 + \int_1^0 v_\theta(x_{\sigma_v'}, c, \sigma_v') \, d\sigma_v'$ with a multistep predictor-corrector solver (FlowUniPC, shift $5.0$) using $6$ steps; the shift concentrates the schedule near $\sigma_v{=}1$ so that the model spends most of its iterations on the high-noise regime. Features are exposed at one chosen step on this $6$-point grid, and we sweep that index over $\{0, 1, 2, 4, 5\}$, corresponding to $\sigma_v \in \{1.000, 0.953, 0.883, 0.562, 0.024\}$.

\paragraph{Action denoising.} The action head is trained with the flow-matching loss $\mathcal{L}_{\mathrm{action}} = \mathbb{E}_{a_0, \varepsilon_a, \sigma_a}\bigl[\lVert v_\phi(a_{\sigma_a}, F_\theta^{(k)}, s, \sigma_a) - (\varepsilon_a - a_0) \rVert^2\bigr]$ under $\sigma_a \sim \mathcal{T}_a$, where $\mathcal{T}_a$ is the beta-distribution $\mathrm{Beta}(1.5, 1.0)\cdot 0.999 + 0.001$ (following~\cite{black2024pi_0}); this favours high action noise during training, matching the inference initial state $a_{\sigma_a{=}1}\sim\mathcal{N}(0,I)$. The target is the constant-conditional velocity $\varepsilon_a - a_0$. At inference we integrate $v_\phi$ from $\sigma_a{=}1$ to $0$ with $10$ Euler steps on a single fixed action chunk of length $A_H$. The video features $F_\theta^{(k)}(x_{\sigma_v},c,\sigma_v)$ are precomputed once per replan and reused across all $10$ Euler steps, so the cost is dominated by the single video pass.

\begin{algorithm}[h]
\caption{Joint Training of $F_\theta$ and $v_\phi$}
\label{alg:training}
\begin{algorithmic}[1]
\Require Extraction layer $k$, flow-time distributions $\mathcal{T}_v, \mathcal{T}_a$, video loss weight $\lambda$
\Repeat
    \State Sample $(x_0, a_0, s, c) \sim \mathcal{D}$ \Comment{video latent, action chunk, state, instruction}
    \State $\sigma_v \sim \mathcal{T}_v$;\quad $\sigma_a \sim \mathcal{T}_a$
    \State $\varepsilon_v, \varepsilon_a \sim \mathcal{N}(0, I)$
    \State $x_{\sigma_v} \gets (1 - \sigma_v)\, x_0 + \sigma_v\, \varepsilon_v$ \Comment{$t{=}0$ frame re-injected as clean}
    \State $a_{\sigma_a} \gets (1 - \sigma_a)\, a_0 + \sigma_a\, \varepsilon_a$
    \State $F^{(k)} \gets F_\theta^{(k)}(x_{\sigma_v}, c, \sigma_v)$ \Comment{forward video DiT up to block $k$}
    \State $\mathcal{L}_{\mathrm{video}} \gets \| v_\theta(x_{\sigma_v}, c, \sigma_v) - (\varepsilon_v - x_0) \|^2$
    \State $\mathcal{L}_{\mathrm{action}} \gets \| v_\phi(a_{\sigma_a}, F^{(k)}, s, \sigma_a) - (\varepsilon_a - a_0) \|^2$
    \State Take gradient step on $\mathcal{L}_{\mathrm{action}} + \lambda\, \mathcal{L}_{\mathrm{video}}$ w.r.t. $\theta, \phi$
\Until{converged}
\end{algorithmic}
\end{algorithm}

\vspace{-0.5em}

Our training procedure is similar to ~\cite{pai2025mimic, ma2026dit4dit}. In two-stage training we instantiate Algorithm~\ref{alg:training} twice with disjoint trainable parameters. \emph{Stage 1} trains only the video backbone by optimising $\mathcal{L}_{\mathrm{video}}$ alone (lines 7--8) with $\phi$ ignored; \emph{Stage 2} freezes $\theta$, caches $F^{(k)}$ at randomly sampled $\sigma_v$, and trains $\phi$ on $\mathcal{L}_{\mathrm{action}}$ alone. Joint training keeps both losses active, with $\lambda{=}1.0$ used throughout.

\begin{algorithm}[h]
\caption{Action Sampling at replan step $i$}
\label{alg:sampling}
\begin{algorithmic}[1]
\Require Extraction layer $k$, target video flow-time $\sigma_v^\star$, action denoising steps $N_a$
\State \textbf{Input:} observation $o_t$, proprioceptive state $s$, language instruction $c$
\State $x_{\sigma_v=1} \sim \mathcal{N}(0, I)$;\quad $a_{\sigma_a=1} \sim \mathcal{N}(0, I)$ \Comment{anchor $t{=}0$ in $x$ replaced by clean $o_t$}
\State $x_{\sigma_v^\star} \gets x_{\sigma_v=1} + \displaystyle\int_{1}^{\sigma_v^\star} v_\theta(x_{\sigma_v'}, c, \sigma_v')\, d\sigma_v'$ \Comment{$\approx$ 6-step FlowUniPC, truncated at $\sigma_v^\star$}
\State $F^{(k)} \gets F_\theta^{(k)}(x_{\sigma_v^\star}, c, \sigma_v^\star)$ \Comment{partial denoising avoids full pixel reconstruction}
\State $a_0 \gets a_{\sigma_a=1} + \displaystyle\int_{1}^{0} v_\phi(a_{\sigma_a}, F^{(k)}, s, \sigma_a)\, d\sigma_a$ \Comment{$N_a{=}10$ Euler steps}
\State \Return action chunk $a_0$
\end{algorithmic}
\end{algorithm}

\vspace{-0.5em}

\paragraph{First-frame anchor mechanism during video denoising.} The $t{=}0$ latent encodes the current image observation $o_t$ and is held \emph{exactly clean} throughout video denoising. The Cosmos-Predict2.5 backbone (\url{https://github.com/nvidia-cosmos/cosmos-predict2.5}) uses three coordinated operations. \textit{(i) Anchor channel:} A binary mask $M \in \{0, 1\}^{B\times 1\times T\times H\times W}$ with $M[:,:,0]{=}1$ and zeros elsewhere is concatenated as an extra channel to the $16$-channel VAE latent, so the DiT receives a $17$-channel input that marks which positions are anchors. \textit{(ii) Pre-network latent replacement:} Before every solver call, the noisy latent is overwritten at the anchor positions with the clean observation latent $x_0^{(t{=}0)}$: $x_{\sigma_v} \leftarrow M \odot x_0^{(t{=}0)} + (1{-}M) \odot x_{\sigma_v}$. The DiT therefore never sees a noised current frame at any $\sigma_v$. \textit{(iii) Post-network velocity replacement:} After the DiT predicts $v_\theta$, the predicted velocity at anchor positions is overwritten with the analytic ``clean'' velocity $\varepsilon_v - x_0^{(t{=}0)}$, so the subsequent Euler/UniPC update lands back on the clean anchor. When configured, the per-position timestep embedding at anchor positions is also replaced with a near-zero constant ($\sigma_v{=}10^{-4}$), which feeds the ``clean'' signal into the adaptive LayerNorm conditioning.

\textbf{Significance for Temporal-Ratio.} The above described asymmetry is what makes TR informative. First frame tokens ($\mathcal{V}_0$) carry a high-SNR, language-independent observation; future frame tokens ($\mathcal{V}_+$) carry the model's noisy imagined rollout, whose SNR is set by $\sigma_v$. The attention the action head puts on $\mathcal{V}_0$ versus $\mathcal{V}_+$ is therefore a routing decision between reacting to the current image and following the imagined future. The monotonic increase of TR as $\sigma_v$ drops, which we report in Sec.~\ref{sec:tr-analysis}, follows directly from holding the anchor clean.

\subsection{TR-Adaptive Guidance Algorithm}
\label{app:pseudocode}

\begin{algorithm}[h]
\caption{TR-Adaptive Velocity-Space Guidance at replan step $i$}
\label{alg:guidance}
\begin{algorithmic}[1]
\Require Observation $o_t$, proprioceptive state $s$, language instruction $c$, target video flow-time $\sigma_v^\star$, action denoising steps $N_a$
\Require Base weights $w_{\mathrm{lang}}^{\mathrm{base}}, w_{\mathrm{plan}}^{\mathrm{base}}$, schedule $\in \{\mathrm{flat}, \mathrm{adaptive}\}$, cap $s_{\max}$
\Require Plan horizon $T' > T$, TR layer $\ell^*$, persistent episode baseline $\mathrm{TR}_0$ (initialised at the first replan, $i{=}0$)
\State \textbf{// Video feature branches (each: 6-step FlowUniPC truncated at $\sigma_v^\star$)}
\State $F^{(k)} \gets F_\theta^{(k)}(x_{\sigma_v^\star}, c, \sigma_v^\star)$ \Comment{conditional}
\If{$w_{\mathrm{lang}}^{\mathrm{base}} > 0$}
    \State $F_{\varnothing}^{(k)} \gets F_\theta^{(k)}(x_{\sigma_v^\star}, \varnothing, \sigma_v^\star)$ \Comment{null-text}
\EndIf
\If{$w_{\mathrm{plan}}^{\mathrm{base}} > 0$}
    \State $F_{\mathrm{plan}}^{(k)} \gets F_\theta^{(k)}(x_{\sigma_v^\star}^{T'}, c, \sigma_v^\star)\big|_{\text{first }T}$ \Comment{extended-horizon, truncated to $T$}
\EndIf
\State
\State \textbf{// Read TR once per replan and update adaptive scale}
\State $\mathrm{TR}_i \gets$ TR at layer $\ell^*$ of $v_\phi(\cdot, F^{(k)}, s, \sigma_a{=}1)$ \Comment{from action-head attention; Eq.~\ref{eq:tr-definition}}
\If{$i = 0$} \quad $\mathrm{TR}_0 \gets \mathrm{TR}_i$ \Comment{episode baseline; reset on episode reset} \EndIf
\If{schedule = flat}
    \State $w_{\mathrm{lang}, i} \gets w_{\mathrm{lang}}^{\mathrm{base}}$, \quad $w_{\mathrm{plan}, i} \gets w_{\mathrm{plan}}^{\mathrm{base}}$
\Else \Comment{TR-adaptive}
    \State $s_i \gets \mathrm{clip}\!\left(\mathrm{TR}_i / \mathrm{TR}_0 - 1,\; 0,\; s_{\max}\right)$
    \State $w_{\mathrm{lang}, i} \gets w_{\mathrm{lang}}^{\mathrm{base}} \cdot s_i$, \quad $w_{\mathrm{plan}, i} \gets w_{\mathrm{plan}}^{\mathrm{base}} \cdot s_i$
\EndIf
\State
\State \textbf{// Velocity-space CFG inside the action Euler loop}
\State $a_{\sigma_a{=}1} \sim \mathcal{N}(0, I)$
\For{Euler step $n = 1, \dots, N_a$ with $\sigma_a: 1 \to 0$}
    \State $v_{\mathrm{cond}} \gets v_\phi(a_{\sigma_a}, F^{(k)}, s, \sigma_a)$
    \If{$w_{\mathrm{lang}, i} > 0$}
        \State $\Delta_{\mathrm{lang}}^v \gets v_{\mathrm{cond}} - v_\phi(a_{\sigma_a}, F_{\varnothing}^{(k)}, s, \sigma_a)$
    \Else
        \State $\Delta_{\mathrm{lang}}^v \gets 0$
    \EndIf
    \If{$w_{\mathrm{plan}, i} > 0$}
        \State $\Delta_{\mathrm{plan}}^v \gets v_\phi(a_{\sigma_a}, F_{\mathrm{plan}}^{(k)}, s, \sigma_a) - v_{\mathrm{cond}}$
    \Else
        \State $\Delta_{\mathrm{plan}}^v \gets 0$
    \EndIf
    \State $v_{\mathrm{guided}} \gets v_{\mathrm{cond}} + w_{\mathrm{lang}, i}\, \Delta_{\mathrm{lang}}^v + w_{\mathrm{plan}, i}\, \Delta_{\mathrm{plan}}^v$
    \State Euler-update $a_{\sigma_a}$ using $v_{\mathrm{guided}}$
\EndFor
\State \Return action chunk $a_0$
\end{algorithmic}
\end{algorithm}

TR-adaptive guidance (Algorithm~\ref{alg:guidance}) reuses the same forward pass as the standard denoising: the conditional velocity $v_\phi(\cdot, F_\theta^{(k)}, s, \sigma_a)$ at the first Euler step is logged through the action head's attention, the TR is read off at layer $\ell^*$, and the per-step weights $w_{\mathrm{lang}}, w_{\mathrm{plan}}$ are rescaled before the remaining Euler steps. Because $a_{\sigma_a{=}1}$ is freshly sampled each replan, the TR baseline $\mathrm{TR}_0$ is reset at $i{=}0$ of every episode, not every replan.

\clearpage

\section{Simulation Task Setup}
\label{app:task_setup}
\label{app:sim_task_setup}

\subsection{LIBERO In-Distribution Suites}
\label{app:libero-id-suites}

\paragraph{Suite definitions.} We use the four standard LIBERO~\cite{Liu2023LIBEROBK} suites: \texttt{libero\_spatial}, \texttt{libero\_object}, \texttt{libero\_goal}, and \texttt{libero\_10} (also \texttt{libero\_long}, for long-horizon tasks). Each suite has $10$ tasks. The action space is $7$-dim end-effector pose ($6$ pose deltas $+ 1$ gripper) and the observation is two $224\times 224$ RGB cameras (\texttt{agentview} and \texttt{eye\_in\_hand}) with an $8$-dim proprioception.

\paragraph{Training demonstrations.} Each task ships with $50$ human demonstrations ($500$ per suite). We use the filtered demonstrations from \url{https://huggingface.co/datasets/physical-intelligence/libero} without any modifications and concatenate the two cameras into the $448\times 224$ input the video backbone expects. The same set trains every paradigm (joint, two-stage, LoRA, full FT).

\paragraph{Evaluation protocol.} We use two evaluation budgets. The headline baseline comparison (Table~\ref{tab:libero_complete}) runs $50$ rollouts per task across $3$ trial seeds: $1{,}500$ per suite, $10{,}500$ total. The design-space grid sweep (Sec.~\ref{sec:guidance}) drops to $10$ rollouts per task at $3$ trial seeds to keep the $60$-cell sweep tractable, giving $700$ rollouts per configuration. Trial seeds hash deterministically over (task, episode-index, seed-id), so reruns reproduce the initial states. The policy runs at $10$Hz (App.~\ref{app:training-inference-procedure}). Max episode length is $220$, $280$, $300$, $520$ steps on \texttt{spatial}, \texttt{object}, \texttt{goal}, \texttt{libero\_10}, and $1.5\times$ those on the OOD counterparts ($330$, $420$, $450$).

\begin{table}[h]
\centering
\caption{LIBERO evaluation suites. Budget shown is for the headline baseline comparison; the grid sweep uses $10$ rollouts per task at $3$ trial seeds.}
\label{tab:app-libero-suites}
\begin{tabular}{llcccc}
\toprule
Suite & Dist. & Tasks & Ep./task & Seeds & Total ep. \\
\midrule
Spatial    & ID  & 10 & 50 & 3 & 1500 \\
Object     & ID  & 10 & 50 & 3 & 1500 \\
Goal       & ID  & 10 & 50 & 3 & 1500 \\
Long       & ID  & 10 & 50 & 3 & 1500 \\
Spatial-OOD & OOD & 10 & 50 & 3 & 1500 \\
Object-OOD  & OOD & 10 & 50 & 3 & 1500 \\
Goal-OOD    & OOD & 10 & 50 & 3 & 1500 \\
\midrule
\multicolumn{2}{l}{Total per config.} & 70 & -- & 3 & 10500 \\
\bottomrule
\end{tabular}
\end{table}

\subsection{LIBERO Compositional OOD Suites}
\label{app:libero-ood-suites}

The three OOD suites are from~\citet{li2025task} \url{https://github.com/QuanyiLi/pi0-text-latent}. They modify the LIBERO BDDL files along one controlled axis at a time. Camera, simulator, robot, and scene match the corresponding ID suite; only the language--object--goal binding moves.

\paragraph{Spatial-OOD (object axis).} Four tasks use new pick targets (\texttt{butter}, \texttt{chocolate\_pudding}, \texttt{milk}, \texttt{orange\_juice}) with the familiar \texttt{plate}. Six use the familiar \texttt{black\_bowl} placed via spatial relations not seen in the ID set (\texttt{on cookie\_box}, \texttt{next\_to plate}, \texttt{at table\_center}, each onto \texttt{stove} or \texttt{cabinet}). The predicate vocabulary (\texttt{on}, \texttt{next\_to}, \texttt{between}) is the same as in training.

\paragraph{Object-OOD (identity axis).} Task names, scene, object positions, and goal predicates match \texttt{libero\_object}; the BDDL only swaps which type is bound to each named instance. For example, \texttt{pick\_up\_the\_butter\_and\_place\_it\_in\_the\_basket.bddl} declares \texttt{butter\_1 - chocolate\_pudding} and \texttt{chocolate\_pudding\_1 - butter}, and the goal binds \texttt{chocolate\_pudding\_1} (the instance that visually looks like butter). A policy that memorised position-to-name fails; only one that grounds language in the current scene succeeds. The swap covers all ten classes.

\paragraph{Goal-OOD (composition axis).} The object inventory matches \texttt{libero\_goal}, but each task pairs an object with a destination it never reaches in ID. The ten pairings include \texttt{wine\_bottle} $\to$ \{\texttt{stove}, \texttt{plate}, \texttt{bowl}\}, \texttt{cream\_cheese} $\to$ \{\texttt{basket}, \texttt{plate}, \texttt{stove}, \texttt{top-of-cabinet}\}, \texttt{orange\_juice} $\to$ \texttt{stove}, \texttt{bbq\_sauce} $\to$ \texttt{plate}, and \texttt{tomato\_sauce} $\to$ \texttt{top-of-cabinet}. Both sub-skills exist in ID tasks; the binding does not. Success requires stitching trajectories from disjoint ID tasks.

\subsection{Relation to Other LIBERO Extensions}
\label{app:libero-extensions-comparison}

Two recent LIBERO extensions overlap with the suites we use. LIBERO-Plus~\cite{fei2025libero} adds $10{,}030$ tasks across seven perturbation dimensions: camera viewpoint, robot pose, lighting, textures, sensor noise, distractors, and paraphrased instructions. LIBERO-PRO~\cite{zhou2025libero} adds five: object appearance/colour/scale, position, paraphrase, task-logic edits, and environment swaps. Both report strong baselines dropping from ${\sim}0.95$ ID success to ${<}0.3$ under their perturbations. LIBERO-PRO's task-logic dimension is the closest analogue of goal-OOD: both replace the goal predicate with a binding absent from training.

The benchmarks are complementary. LIBERO-Plus and LIBERO-PRO sweep many dimensions in one benchmark, mixing perceptual and compositional shifts while LIBERO-OOD's three suites all sit on the language--object--goal binding axis. The Temporal Ratio analysis of Sec.~\ref{sec:tr-analysis} probes that axis, which is why we use LIBERO-OOD here.


\section{Design-Space Analysis and Additional Results}
\label{app:sim_design_analysis}

\subsection{Full Training, Noise-Level, and Horizon Sweep}
\label{app:full-design-sweep}

\paragraph{Finetuning strategy.} LoRA dominates Full FT on OOD across every cell of the sweep. Joint $+$ LoRA peaks at $55.7$\% OOD (Spatial peak $\sigma_v{=}0.883$, $T{=}5$); Two-Stage $+$ LoRA peaks at $51.0$\% OOD. Full FT show competitive ID performance (Two-Stage $+$ Full reaches $95.2$\% ID at $\sigma_v{=}0.883$, $T{=}5$) but lacks OOD generalization ($33.0$\% OOD SR). The pattern is consistent with full-DiT updates destroying the pretrained video prior that supports OOD recomposition.

\paragraph{Video noise level.} For Joint $+$ LoRA at $T{=}5$, both ID and OOD peak at the same intermediate noise level $\sigma_v{=}0.883$ ($94.3$\% ID, $55.7$\% OOD); pure noise ($\sigma_v{=}1.0$) costs $16$\,pp on ID and $15$\,pp on OOD, and the near-clean extraction ($\sigma_v{=}0.024$) costs another $12$\,pp on ID. The non-monotonic pattern repeats for the other paradigms: every $T{=}5$ row hits its joint ID/OOD optimum at $\sigma_v \in \{0.883, 0.953\}$. Two-Stage models are more robust to noise level and maintain performance at low video noise levels ($\sigma_v{=}0.024$).

\paragraph{Prediction horizon.} The horizon ablation only changes how far the video DiT rolls out at evaluation time. \emph{Training is always done with a $T{=}5$ latent-frame prediction target}, and at evaluation we always pass the first $5$ latent frames of the video rollout to the action head, irrespective of weight video denoising happens for $5$, $9$, or $24$ latent frames frames. Increasing the imagined rollout horizon therefore only changes the quality of the first $5$ latent frames (longer video horizon allows the video model to maintain object-goal relationships), not the number of frames seen by the action head. With that setup, $T{=}5$ at evaluation matches the training distribution and dominates every paradigm at every $\sigma_v$. Stretching the evaluation rollout to $T{=}9$ costs $20$--$40$\,pp on most cells; $T{=}24$ collapses Joint variants almost entirely. Similar to video-noise level study, Two-Stage models do not hurt the video prior at all and helps in maintaining imagined rollout's physical plausibility. The asymmetric collapse mirrors the failure mode in Sec.~\ref{sec:guidance}: while longer rollouts preserve the object-goal relationship, they incorporate more hallucinations like frame jumping and implausible transitions.

\begin{table}[h]
\centering
\caption{Baseline SR (\%) by training paradigm, denoising target step, and horizon on LIBERO. LoRA rows use $r{=}32$.}
\label{tab:paradigm-sigma-horizon}
\small
\setlength{\tabcolsep}{4pt}
\begin{tabular}{ll cc cc cc cc cc}
\toprule
& & \multicolumn{2}{c}{$\sigma{\approx}1.0$} & \multicolumn{2}{c}{$\sigma{\approx}0.953$} & \multicolumn{2}{c}{$\sigma{\approx}0.883$} & \multicolumn{2}{c}{$\sigma{\approx}0.562$} & \multicolumn{2}{c}{$\sigma{\approx}0.024$} \\
\cmidrule(lr){3-4} \cmidrule(lr){5-6} \cmidrule(lr){7-8} \cmidrule(lr){9-10} \cmidrule(lr){11-12}
Paradigm & Horizon & ID & OOD & ID & OOD & ID & OOD & ID & OOD & ID & OOD \\
\midrule
  \multirow{3}{*}{Joint + LoRA}
      & $T{=}5$  & 78.0 & 41.0 & 88.0 & 53.7 & 94.3 & 55.7 & 83.5 & 49.7 & 82.5 & 45.3 \\
      & $T{=}9$  & 69.5 & 34.3 & 56.2 & 40.3 & 49.2 & 32.0 & 49.5 & 28.3 & 43.2 & 25.0 \\
      & $T{=}24$ & 13.8 & 11.3 & 5.8  & 9.7  & 0.0  & 2.7  & 0.2  & 2.3  & 0.0  & 1.0  \\
  \midrule
  \multirow{3}{*}{Joint + Full}
      & $T{=}5$  & 72.0 & 23.0 & 88.5 & 29.3 & 93.7 & 33.0 & 88.2 & 24.7 & 82.5 & 18.7 \\
      & $T{=}9$  & 42.0 & 8.3  & 47.0 & 11.3 & 42.0 & 14.0 & 45.0 & 11.3 & 33.8 & 6.7  \\
      & $T{=}24$ & 3.0  & 0.3  & 7.5  & 5.3  & 3.5  & 3.7  & 1.2  & 1.3  & 1.0  & 1.0  \\
  \midrule
  \multirow{3}{*}{Two-Stage + LoRA}
      & $T{=}5$  & 88.0 & 38.7 & 95.0 & 51.0 & 96.5 & 46.0 & 95.8 & 40.7 & 92.2 & 42.3 \\
      & $T{=}9$  & 61.8 & 22.0 & 71.2 & 37.0 & 69.0 & 39.7 & 71.2 & 37.3 & 71.5 & 41.3 \\
      & $T{=}24$ & 13.5 & 10.0 & 28.2 & 26.7 & 22.0 & 25.7 & 22.8 & 26.3 & 21.0 & 28.7 \\
  \midrule
  \multirow{3}{*}{Two-Stage + Full}
      & $T{=}5$  & 90.0 & 11.3 & 92.5 & 22.0 & 95.2 & 29.7 & 94.2 & 29.0 & 93.0 & 29.7 \\
      & $T{=}9$  & 78.0 & 4.7  & 78.2 & 10.3 & 74.8 & 13.3 & 64.2 & 17.0 & 58.5 & 15.0 \\
      & $T{=}24$ & 63.5 & 1.7  & 48.5 & 7.0  & 33.5 & 11.7 & 20.2 & 12.7 & 16.2 & 12.7 \\
  \bottomrule
\end{tabular}
\end{table}

\subsection{Per-Suite Breakdown}
\label{app:per-suite}

\begin{table}[h]
\centering
\caption{Per-suite ID SR (\%) for the unguided baseline, flat guidance, and TR-adaptive guidance. Combined guidance uses $w_{\mathrm{lang}}{=}0.1$, $w_{\mathrm{plan}}{=}0.3$, plan horizon $T'{=}24$.}
\label{tab:app-persuite-id}
\begin{tabular}{ll cccc c}
\toprule
Configuration & Setting & Spatial & Object & Goal & Long & Avg ID \\
\midrule
Ours (unguided)        & $\sigma_v{=}0.883$, $T{=}5$ & $97.3$ & $98.0$ & $97.6$ & $84.3$ & $94.3$ \\
Ours (flat guided)     & combined, flat schedule     & $85.6$ & $90.6$ & $88.3$ & $72.3$ & $84.3$ \\
Ours (adaptive guided) & combined, TR-adaptive       & $96.3$ & $99.6$ & $97.6$ & $82.6$ & $94.0$ \\
\bottomrule
\end{tabular}
\end{table}

\begin{table}[h]
\centering
\caption{Per-suite OOD SR (\%) for the unguided baseline, flat guidance, and TR-adaptive guidance}
\label{tab:app-persuite-ood}
\begin{tabular}{ll ccc c}
\toprule
Configuration & Setting & Spatial & Object & Goal & Avg OOD \\
\midrule
Ours (unguided)        & $\sigma_v{=}0.883$, $T{=}5$ & $50.3$ & $40.0$ & $77.3$ & $55.7$ \\
Ours (flat guided)     & combined, flat schedule     & $53.6$ & $40.0$ & $78.3$ & $57.3$ \\
Ours (adaptive guided) & combined, TR-adaptive       & $58.6$ & $40.0$ & $80.3$ & $59.4$ \\
\bottomrule
\end{tabular}
\end{table}

\paragraph{Flat versus TR-adaptive scheduling.} Both schedules use the same combined guidance signal ($w_{\mathrm{lang}}{=}0.1$, $w_{\mathrm{plan}}{=}0.3$); they differ only in how the per-step weight is set. Flat applies the base weight at every action Euler step; TR-adaptive multiplies it by $s_i = \mathrm{clip}(\mathrm{TR}_i / \mathrm{TR}_0 - 1, 0, s_{\max})$, so the guidance weight increases only when TR rises above its episode baseline (Alg.~\ref{alg:guidance}). While both schedules help in improving OOD success by employing more compositional video guidance from the backbone, flat guidance applies guidance even during the precise manipulation phases. This makes grasping inaccurate and flat-guidance leads to a significant loss in ID performance. The TR analysis of Sec.~\ref{sec:tr-analysis} predicts this split: planning-heavy phases of an episode have high TR and benefit from amplified video conditioning, while precise-manipulation phases have low TR and need to react to the current frame. TR-adaptive scheduling concentrates the guidance into the planning phases without interfering with the precise manipulation phases.

\subsection{Video Feature Extraction Layer Ablation}
\label{app:layer-ablation}

\begin{figure}[h]
\centering
\includegraphics[width=\linewidth]{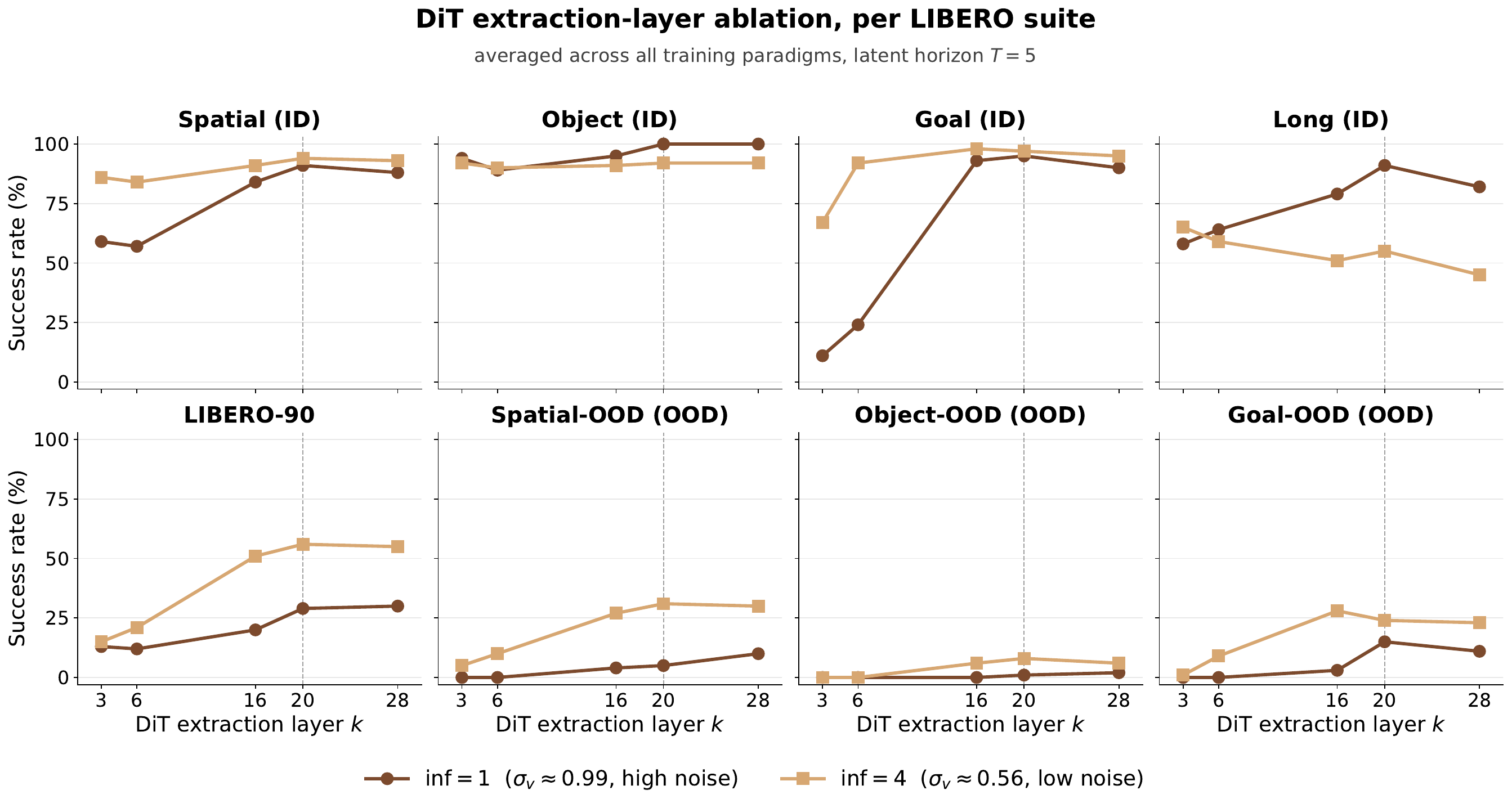}
\caption{DiT extraction layer ablation across the seven LIBERO suites plus LIBERO-$90$, averaged across the four training paradigms at $T{=}5$. The two noise levels probe the same layer grid: inf$=$1 ($\sigma_v{\approx}0.99$, high noise) and inf$=$4 ($\sigma_v{\approx}0.562$, low noise). $k{=}20$ is the joint optimum used throughout the rest of the paper.}
\label{fig:app-layer-ablation}
\end{figure}

\paragraph{DiT block selection.} We sweep the extraction layer over $k \in \{3, 6, 16, 20, 28\}$ ($28$-block DiT, $1$-indexed) under two noise levels: a high-noise extraction (inf$=$1, $\sigma_v{\approx}0.99$) and a low-noise extraction (inf$=$4, $\sigma_v{\approx}0.562$). Each cell is averaged across the four training paradigms (Joint $\times$ LoRA/Full, Two-Stage $\times$ LoRA/Full) at latent horizon $T{=}5$. Results are summarised in Table~\ref{tab:app-layer-ablation} and per-suite breakdown is provided in~\autoref{fig:app-layer-ablation}. $k{=}20$ is the joint optimum across ID and OOD and aligns with the findings of ~\cite{pai2025mimic, ma2026dit4dit}; layers shallower than $k{=}16$ are clearly suboptimal under both noise levels.

\begin{table}[h]
\centering
\caption{Effect of DiT extraction layer $k$ on Avg ID / Avg OOD SR (\%). Averaged across the four training paradigms at latent horizon $T{=}5$. inf$=$1 corresponds to the high-noise regime ($\sigma_v{\approx}0.99$), inf$=$4 to the low-noise regime ($\sigma_v{\approx}0.562$).}
\label{tab:app-layer-ablation}
\small
\setlength{\tabcolsep}{4pt}
\begin{tabular}{l cc cc cc cc cc}
\toprule
& \multicolumn{2}{c}{$k{=}3$} & \multicolumn{2}{c}{$k{=}6$} & \multicolumn{2}{c}{$k{=}16$} & \multicolumn{2}{c}{$k{=}20$} & \multicolumn{2}{c}{$k{=}28$} \\
\cmidrule(lr){2-3} \cmidrule(lr){4-5} \cmidrule(lr){6-7} \cmidrule(lr){8-9} \cmidrule(lr){10-11}
Noise level & ID & OOD & ID & OOD & ID & OOD & ID & OOD & ID & OOD \\
\midrule
inf$=$1 ($\sigma_v{\approx}0.99$) & $55.5$ & $0.0$ & $58.5$ & $0.0$ & $87.8$ & $2.3$ & $\mathbf{94.3}$ & $7.0$  & $90.0$ & $7.7$  \\
inf$=$4 ($\sigma_v{\approx}0.562$) & $77.5$ & $2.0$ & $81.3$ & $6.3$ & $82.8$ & $20.3$ & $84.5$ & $\mathbf{21.0}$ & $81.3$ & $19.7$ \\
\bottomrule
\end{tabular}
\end{table}

\subsection{Gemma Layer selection for TR}
\label{app:tr-extended}

\begin{figure}[h]
\centering
\includegraphics[width=0.6\linewidth]{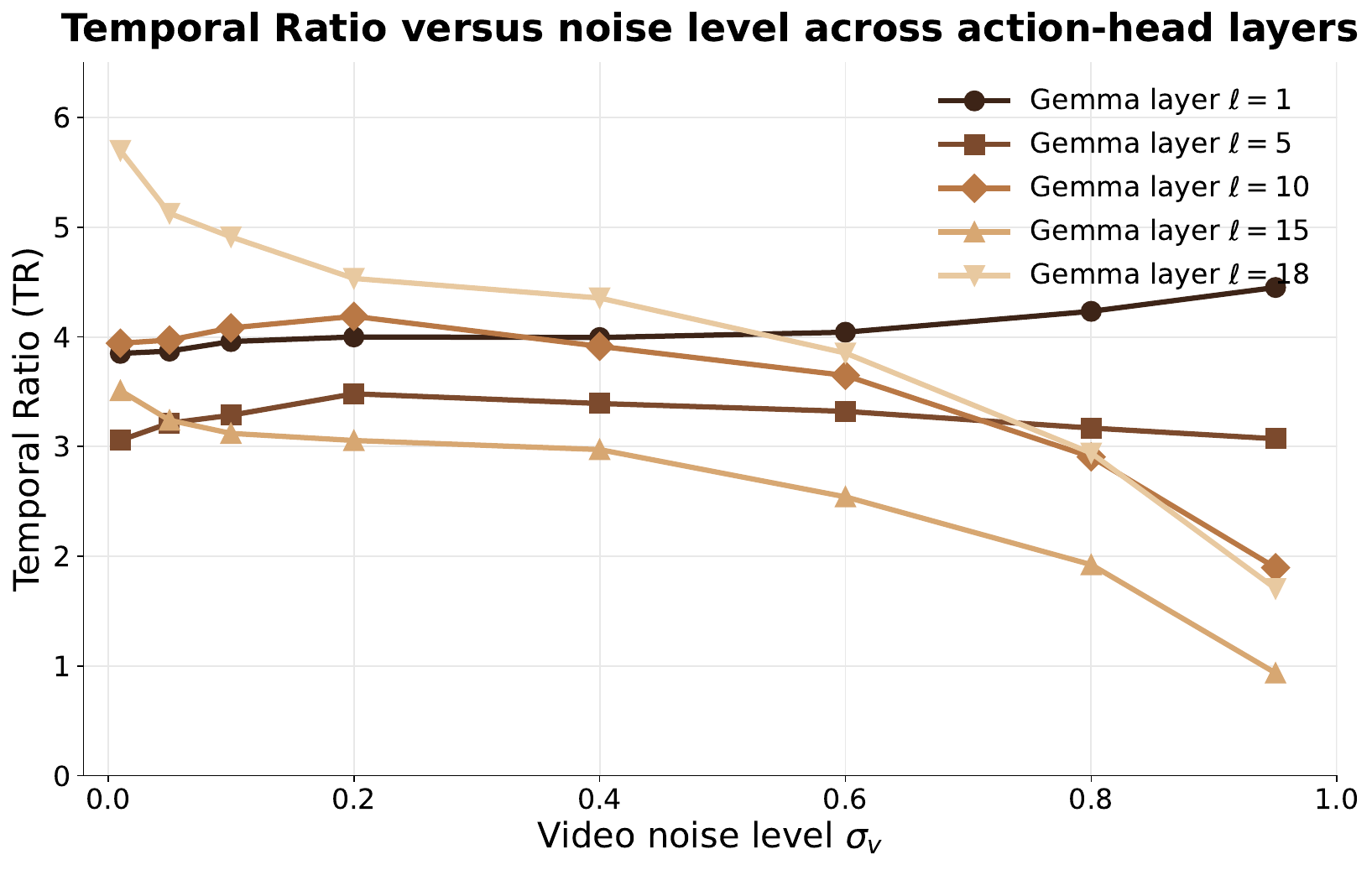}
\caption{Temporal Ratio as a function of video noise level $\sigma_v$ at five layers of the $18$-layer Gemma action head (1-indexed). Shallow layers are insensitive to $\sigma_v$; deep layers (and especially $\ell{=}15, 18$) carry the routing decision between the clean current frame and the noisy predicted futures. We pick $\ell^* = 15$ as the canonical TR layer for adaptive guidance and reported TR values.}
\label{fig:app-tr-layers}
\end{figure}

We choose Gemma layer $\ell^* = 15$ for all reported TR values. The choice depends on how much each layer's attention split between the current frame and the predicted future frames moves as the extraction noise level $\sigma_v$ changes. Fig.~\ref{fig:app-tr-layers} plots TR against $\sigma_v$ at five layers spanning the $18$-block Gemma expert. Early layers ($\ell{=}1, 5$) sit near a fixed value (TR ${\approx}\,4$ and ${\approx}\,3.2$ respectively) across the entire $\sigma_v$ range indicating equal importance to all frames (1 current vs 4 future frames). For deep layers ($\ell{=}15, 18$): at low $\sigma_v$ (clean futures) TR is high, and as $\sigma_v$ rises (noisier futures) TR collapses toward $1$ or below, because the action head reroutes its attention onto the clean current-frame anchor. The mid-layer $\ell{=}10$ shows the same pattern with a smaller range, so the routing decision sharpens with depth.

\section{Implementation and Hyperparameter Details}
\label{app:implementation_details}

\subsection{Video Backbone Adaptation}
\label{app:video-hparams}

\paragraph{LoRA adaptation.} We adapt the DiT backbone (App.~\ref{app:latent-vam-architecture}) with Low-Rank Adaptation~\cite{hu2022lora} at ranks $r \in \{16, 32, 64\}$ and $\alpha = r$. LoRA is inserted into every attention and MLP projection in all $28$ blocks, which trains $2$--$6\%$ of the backbone parameters (${\sim}40$M at $r{=}16$, ${\sim}180$M at $r{=}64$); the rest of the DiT is frozen.

\paragraph{Full finetuning.} We also evaluate updating all $2.1$B DiT parameters. Optimizer settings match the LoRA configuration except for the learning rate, which is lowered from $10^{-4}$ to $10^{-5}$ to avoid destabilising the pretrained features.

\paragraph{Joint and two-stage training.} Joint training (Alg.~\ref{alg:training}) runs for $50$k steps with video loss weight $\lambda = 1.0$. Two-stage training runs Stage 1 (video only) for $25$k steps and Stage 2 (action head only, video features cached) for $50$k steps. Both modes share the optimizer, batch size, and noise schedule below.

\begin{table}[h]
\centering
\caption{Video model finetuning hyperparameters.}
\label{tab:app-video-hparams}
\begin{tabular}{lcc}
\toprule
Hyperparameter & LoRA & Full FT \\
\midrule
Learning rate           & $1 \times 10^{-4}$ & $1 \times 10^{-5}$ \\
LoRA rank $r$           & 32   & -- \\
LoRA $\alpha$           & $\alpha = r$        & -- \\
LoRA targets            & all $Q/K/V/O$ and MLP projections, $28$ blocks & -- \\
Trainable params        & $2$--$6\%$ of $2.1$B   & ${\sim}2.1$B ($100\%$) \\
Optimizer               & AdamW ($\beta_1{=}0.9, \beta_2{=}0.95, \varepsilon{=}10^{-8}$) & same \\
Weight decay            & $1 \times 10^{-8}$  & $1 \times 10^{-8}$ \\
Gradient clip (norm)    & $1.0$               & $1.0$ \\
Training steps          & $50$k joint, $25$k Stage 1 & $50$k joint \\
Batch size              & $8$ per GPU $\times$ $16$ GPUs $=$ $128$ & same \\
LR schedule             & cosine decay to $5\times 10^{-7}$ & same \\
EMA decay               & $0.999$ (constant) & $0.999$ \\
Video noise schedule    & $\sigma_v \sim \mathcal{U}[0, 1]$ & $\sigma_v \sim \mathcal{U}[0, 1]$ \\
Joint loss weight $\lambda$ & $1.0$           & $1.0$ \\
\bottomrule
\end{tabular}
\end{table}

\subsection{Action Head Training}
\label{app:action-hparams}

\paragraph{Action flow matching.} The action flow-matching loss and Euler-step inference procedure are defined in App.~\ref{app:training-inference-procedure}. The empirical setting we use is $\sigma_a \sim \mathrm{Beta}(1.5, 1.0) \cdot 0.999 + 0.001$~\cite{intelligence2025pi}, which biases the training schedule toward high noise and matches the $\sigma_a{=}1$ initialisation at inference.

\paragraph{Action chunking.} Action chunks are $17$ steps and the policy replans every $10$ environment steps (App.~\ref{app:training-inference-procedure}). The action vector has $7$ dimensions per arm on LIBERO (six joint deltas plus a binary gripper) and is padded to internal dimension $32$, so the same network serves bimanual setups without architecture changes.

\begin{table}[h]
\centering
\caption{Action head training hyperparameters.}
\label{tab:app-action-hparams}
\begin{tabular}{lc}
\toprule
Hyperparameter & Value \\
\midrule
Architecture          & Gemma expert (App.~\ref{app:latent-vam-architecture}; $18$ layers, ${\sim}300$M) \\
Action prediction     & flow matching (single-step training, $10$-step Euler inference) \\
Action chunk size     & $17$ steps \\
Action dim            & $7$ per arm (LIBERO); padded to internal dim $32$ \\
Proprioceptive dim    & $8$, padded to internal dim $32$ \\
Action noise schedule & $\sigma_a \sim \mathrm{Beta}(1.5, 1.0) \cdot 0.999 + 0.001$ \\
Learning rate         & $1 \times 10^{-4}$ \\
Training steps (joint and two-stage) & $50$k \\
Optimizer             & AdamW ($\beta_1{=}0.9, \beta_2{=}0.95, \varepsilon{=}10^{-8}$) \\
Weight decay          & $1 \times 10^{-8}$ \\
Gradient clip (norm)  & $1.0$ \\
\bottomrule
\end{tabular}
\end{table}

\subsection{Feature Extraction}
\label{app:feature-extraction}

\paragraph{Extraction layer and video noise level.} We read intermediate features from block $k = 20$ of $28$ in the DiT, matching the choice of~\cite{pai2025mimic}. The token shape, patchify scheme, and projection into the action head are described in App.~\ref{app:latent-vam-architecture}. The $6$-step FlowUniPC schedule (App.~\ref{app:training-inference-procedure}) gives a grid of six $\sigma_v$ values. We sweep the extraction noise level across all values covering pure noise, near-noise, the mid-noise regime, and the near-clean regime. 

\paragraph{Prediction horizon.} Three temporal extents are evaluated at inference: $T \in \{5, 9, 24\}$ latent frames, corresponding to $17$, $33$, and $93$ video frames at the $4\times$ temporal compression. Training is always done with a $T{=}5$ target. The horizon parameter only controls how many latent frames the video DiT rolls out during evaluation; the action head always receives the \emph{first} $5$ latent frames of the rollout regardless of $T$, so a larger $T$ changes the quality of those first $5$ frames (joint denoising over a longer chunk) rather than the number of frames the action head sees. $T = 5$ at inference matches training; $T = 24$ amounts to roughly $9$ seconds of imagined future at $10$Hz control. Plan guidance (App.~\ref{app:guidance-hparams}) selects $T'$ from this same set.

\begin{table}[h]
\centering
\caption{Feature extraction configurations.}
\label{tab:app-extraction}
\begin{tabular}{lc}
\toprule
Parameter & Values \\
\midrule
Extraction layer $k$            & $20$ (of $28$ DiT blocks) \\
Latent temporal frames $T$       & $\{5, 9, 24\}$ ($= 17, 33, 93$ video frames) \\
Spatial latent resolution $W \times H$ & $28 \times 14$ (from $448 \times 224$ input) \\
Feature dim $D$                  & $2048$ \\
Video tokens per latent frame    & $98$ (after $1\times 2\times 2$ patchify) \\
Video tokens fed to action head  & $98 T$ \\
$\sigma_v$ values (inf step $\to$ $\sigma_v$) & $0{:}1.000,\ 1{:}0.953,\ 2{:}0.883,\ 4{:}0.562,\ 5{:}0.024$ \\
Video solver steps               & $6$ (FlowUniPC, shift $5.0$) \\
Action Euler steps               & $10$ \\
\bottomrule
\end{tabular}
\end{table}

\subsection{Guidance Hyperparameters}
\label{app:guidance-hparams}

\paragraph{Language guidance.} We compute the null-text branch by replacing the text conditioning $c$ with the zero embedding before the video forward pass. The resulting velocity difference $v_{\mathrm{cond}} - v_{\mathrm{uncond}}$ is scaled by $w_{\mathrm{lang}, i}$ at each action Euler step (Algorithm~\ref{alg:guidance}). The sweep covers $w_{\mathrm{lang}}^{\mathrm{base}} \in \{0.04, 0.06, 0.1, 0.2, 0.3, 0.5, 1.0\}$; the canonical setting in the main results is $w_{\mathrm{lang}}^{\mathrm{base}} = 0.1$.

\paragraph{Plan guidance.} The plan branch queries the video model with the extended horizon $T' = 24$ latent frames ($93$ video frames at $10$Hz, roughly $9$ seconds of imagined future) and truncates the resulting features back to the policy horizon $T$ before passing them to the action head. The velocity difference $v_{\mathrm{plan}} - v_{\mathrm{cond}}$ is scaled by $w_{\mathrm{plan}, i}$. The sweep covers $w_{\mathrm{plan}}^{\mathrm{base}} \in \{0.04, 0.06, 0.1, 0.2, 0.3, 0.5, 1.0\}$ with canonical $w_{\mathrm{plan}}^{\mathrm{base}} = 0.3$.

\paragraph{TR-adaptive schedule.} Adaptive scaling reads $\mathrm{TR}_i$ from layer $\ell^* = 15$ of the action head's attention. The episode baseline $\mathrm{TR}_0$ is set at the first replan of each episode and held until the episode terminates. The scale $s_i = \mathrm{clip}(\mathrm{TR}_i / \mathrm{TR}_0 - 1, 0, s_{\max})$ uses $s_{\max} = 2.0$.

\begin{table}[h]
\centering
\caption{Guidance hyperparameters.}
\label{tab:app-guidance-hparams}
\begin{tabular}{lc}
\toprule
Hyperparameter & Values \\
\midrule
$w_{\mathrm{lang}}^{\mathrm{base}}$ & $0.1$ \\
$w_{\mathrm{plan}}^{\mathrm{base}}$ & $0.3$ \\
Planning horizon $T'$ & $\{24\}$ latent frames \\
Adaptive cap $s_{\max}$ & $2.0$ \\
TR extraction layer $\ell^*$ & $15$ \\
Schedules & flat, TR-adaptive \\
\bottomrule
\end{tabular}
\end{table}


\section{Real-World Setup and Evaluation Details}
\label{app:real_world_setup}

\subsection{Hardware Description}
\label{app:hw-description}

\begin{figure}[h]
\centering
\includegraphics[width=0.75\linewidth]{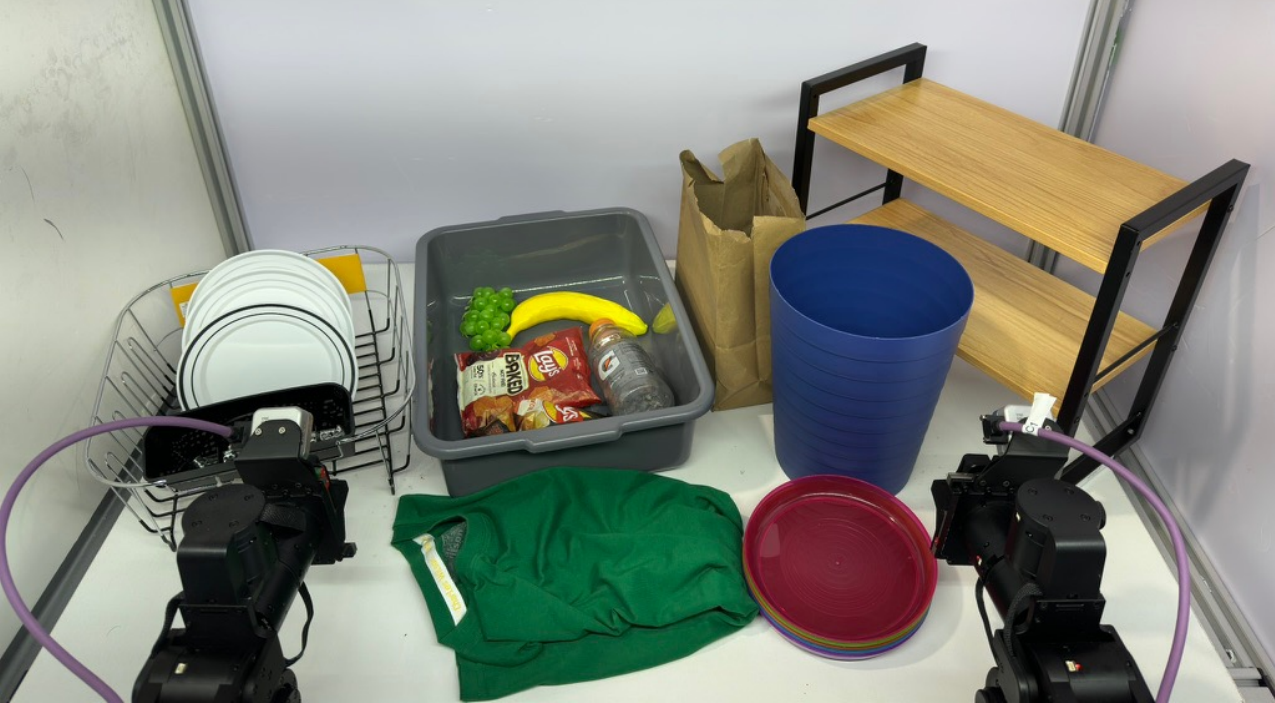}
\caption{\textbf{Bimanual real-world testbed.} Two I2RT YAM 6-DOF arms~\cite{i2rt_yam_arm} and three Intel RealSense D405: one mounted top-down on the workspace and two more mounted on the wrists. The two arms together expose $12$ joint DoF plus $2$ gripper channels, which maps to the $14$-dim action used by the policy ($7$ per arm). We use the RGB stream from three cameras where the native $1280 \times 720$ output is downsampled to $224 \times 168$ before being concatenated into the policy input.}
\label{fig:app-real-world-setup}
\end{figure}

\subsection{Bimanual YAM Platform}
\label{app:yam-platform}

\paragraph{Observation space.} The policy reads three RGB cameras per timestep: a top-mounted workspace view, a left-wrist camera, and a right-wrist camera. Each camera produces a $224 \times 168$ frame; the three are concatenated horizontally into a single $672 \times 168$ image before input to the video backbone (the same flat-image convention used for LIBERO; see App.~\ref{app:libero-id-suites}). Cameras run at $30$Hz; the policy applies a frame-skip of $3$, so the $17$ video frames it sees per replan span $51$ source frames and form a $10$Hz $1.7$-second window. After the Cosmos-Tokenizer $4\times$ temporal compression these $17$ video frames yield $T = 5$ latent frames, matching the simulation configuration. In addition we also use the absolute joint angles for each inference step.

\paragraph{Action space.} The action is $14$-dim absolute joint angle targets, organised as $7$ dimensions per arm. The flow-matching head from simulation (App.~\ref{app:action-hparams}) is reused without architectural change; only the action width and proprioceptive width differ. Internally the $14$-dim action is zero-padded to the action head's internal width of $32$. Each replan produces a chunk of $50$ actions at $30$Hz (a $1.67$-second open-loop horizon); the first $30$ actions ($1$ second) are executed and the rest are discarded before the next replan.

\paragraph{Latent token count.} The $672 \times 168$ stacked image, after $8\times$ spatial compression by Cosmos-Tokenizer-CV4x8x8, yields an $84 \times 21$ latent grid per frame. The DiT $1 \times 2 \times 2$ patchifies, pads  and groups spatial neighbours into $42 \times 11 = 462$ DiT tokens per latent frame, so the full $T = 5$ video prefix presented to the action head contains $5 \times 462 = 2310$ tokens. For reference, the corresponding LIBERO numbers are $98$ tokens per latent frame and $5 \times 98 = 490$ total tokens at $T = 5$.

\subsection{Real-World Training Dataset}
\label{app:real-world-training-dataset}

\paragraph{Dataset.} Real-world training uses a multi-task bimanual YAM dataset consisting of $24$ tasks (a selected subset from~\cite{abc2026}). Overall, the selected dataset contains $5{,}600$ episodes for a total of $9.98\mathrm{M}$ raw camera frames at $30$~fps. (${\sim}3.3\mathrm{M}$ after frame-skipping at $10$~fps). The natural-language instructions are paraphrased per episode: the dataset has $1{,}036$ unique prompt strings spread across the $24$ task categories, so the policy sees a wider distribution of instruction wordings.

\paragraph{Task families.} The $24$ tasks fall into four skill categories: \emph{folding} (1 task), \emph{garment manipulation} (2 tasks; removing shirts and t-shirts from hangers), \emph{high-precision assembly} (1 task), and \emph{pick-and-place} (20 tasks: bussing tables, loading and unloading dish racks, placing snacks and beverages on shelves, throwing items into bins, handoffs, mug-orientation correction, and a relative-placement task). Per-task episode counts are listed in Table~\ref{tab:app-pp24-tasks}.

\paragraph{Sampling weights.} During training the action head sees a \texttt{ConcatDataset} of all $24$ task repositories with a \texttt{WeightedRandomSampler} that assigns equal weight to each of the 24 tasks. 

\begin{table}[h]
\centering
\caption{Multi-task dataset task list, organised by skill family. Episode counts sum to $5{,}600$.}
\label{tab:app-pp24-tasks}
\small
\begin{tabular}{lll}
\toprule
Skill family & Task & Episodes \\
\midrule
folding              & Fold a pile of t-shirts and stack them                & 200 \\
\midrule
\multirow{2}{*}{garment manipulation}
                     & Remove the shirt from the hanger                        & 200 \\
                     & Remove the t-shirt from the hanger                      & 200 \\
\midrule
high-precision assembly & Unscrew bottle caps                                  & 200 \\
\midrule
\multirow{20}{*}{pick-and-place}
                     & Place mixed dishes and glasses into a plastic bin     & 200 \\
                     & Place plates into a plastic bin           & 200 \\
                     & Bimanual object handoff                                  & 200 \\
                     & Load bowls into a tabletop dish rack                    & 200 \\
                     & Load cups into a tabletop dish rack                     & 200 \\
                     & Load mixed dishes into a tabletop dish rack             & 200 \\
                     & Load plates into a tabletop dish rack                   & 200 \\
                     & Place and organize beverages on a shelf                 & 200 \\
                     & Place and organize canned foods on a counter            & 200 \\
                     & Place and organize chip bags on a shelf                 & 200 \\
                     & Place fake fruits into a fruit bowl                     & 200 \\
                     & Place snacks into a paper bag                           & 200 \\
                     & Place an object at a specified relative position        & 1000 \\
                     & Take snacks out of a paper bag                          & 200 \\
                     & Throw plastic bottles into a bin                        & 200 \\
                     & Turn a mug right-side up                                & 200 \\
                     & Unload bowls from a tabletop dish rack                  & 200 \\
                     & Unload cups from a tabletop dish rack                   & 200 \\
                     & Unload mixed dishes from a tabletop dish rack           & 200 \\
                     & Unload plates from a tabletop dish rack                 & 200 \\
\bottomrule
\end{tabular}
\end{table}

\subsection{Real-World Compositional Evaluation}
\label{app:real-world-compositional-evaluation}

\paragraph{Evaluation scenes.} All training scenarios had only one target receptacle and multiple instances of the same target object. We construct three OOD scenes to evaluate compositional generalisation. Each scene contains multiple target candidates and multiple receptacle candidates. \textbf{Scene 1} places a paper bag and a bin together with fruits, testing whether the policy can bind the named target object to the correct receptacle. \textbf{Scene 2} mixes snacks and bottles against the bag and bin, testing the same binding under a multi-object multi-receptacle shift. \textbf{Scene 3 } presents snacks together with a bin and a shelf, testing whether the policy overfits to the \texttt{snack-on-shelf} behaviour seen frequently during training.

\paragraph{Task definitions and trials.} Each scene contains a fixed set of language instructions. The eight evaluation tasks are listed in Table~\ref{tab:app-realworld-tasks}. Tasks 1 and 2 are evaluated for $15$ trials each ($5$ trials with $3$ fruits each); tasks 3 through 8 are evaluated for $5$ trials each. Total budget per method: $60$ trials. A trial succeeds when the robot approaches and picks up the named target object first and reaches the named receptacle. 

\begin{table}[h]
\centering
\caption{Real-world evaluation tasks, organised by scene.}
\label{tab:app-realworld-tasks}
\small
\begin{tabular}{clcc}
\toprule
Task & Instruction & Scene & Trials \\
\midrule
1 & Throw away the fruits into the bin                   & 1 & 15 \\
2 & Place the fruits into the paper bag                  & 1 & 15 \\
3 & Place snacks into the paper bag                      & 2 & 5  \\
4 & Throw away the snacks into the bin                   & 2 & 5  \\
5 & Place plastic bottle into the paper bag              & 2 & 5  \\
6 & Throw away the plastic bottle into the bin           & 2 & 5  \\
7 & Place snacks into the bin                            & 3 & 5  \\
8 & Arrange snacks on the shelf                          & 3 & 5  \\
\midrule
\multicolumn{3}{l}{Total trials per method}              & 60 \\
\bottomrule
\end{tabular}
\end{table}

\subsection{Real-World Training Hyperparameters and Compute}
\label{app:hardware-training}

\paragraph{Optimisation.} We deploy joint LoRA training only in the real-world and use the same optimiser configuration as the simulation runs (App.~\ref{app:video-hparams}, App.~\ref{app:action-hparams}). The video DiT is adapted with LoRA at $r = 32$, $\alpha = 32$, learning rate $1 \times 10^{-4}$. The action head trains at peak learning rate $5 \times 10^{-4}$ with $500$ warmup steps and cosine decay to $5 \times 10^{-5}$ over $10{,}000$ steps. EMA decay is $0.999$, weight decay $1 \times 10^{-8}$, gradient clip $1.0$. Precision is bfloat16.

\paragraph{Baselines.} We compare against $\pi_0$~\cite{black2024pi_0}, $\pi_{0.5}$~\cite{intelligence2025pi}, and Cosmos-Policy~\cite{kim2026cosmos}, all retrained on the same dataset. All WAM and VAM baselines use the same Cosmos-Predict 2.5-2B video backbone as our model; the $\pi$-family baselines are at a comparable $3$B PaliGemma VLM scale.

\paragraph{Compute budget.} Each method is trained for $1\mathrm{M}$ optimiser steps on an equivalent of $32$ NVIDIA H200 80GB GPUs at matched effective batch size and bfloat16 precision.

\begin{table}[h]
\centering
\caption{Real-world training compute and key hyperparameters.}
\label{tab:app-training-hardware}
\begin{tabular}{lc}
\toprule
Parameter & Value \\
\midrule
GPU type                              & NVIDIA H200 80GB \\
Number of GPUs                         & $32$ \\
Effective batch size                    & 512 \\
Training steps                         & $1\mathrm{M}$ \\
Precision                              & bfloat16 \\
LoRA rank $r$ / $\alpha$               & $32$ / $32$ \\
LoRA learning rate                     & $1 \times 10^{-4}$ \\
Action-head peak LR     & $5 \times 10^{-4}$  \\
EMA decay                              & $0.999$ \\
\bottomrule
\end{tabular}
\end{table}

\subsection{Inference Cost}
\label{app:hardware-inference}

\paragraph{Forward-pass cost.} The unguided policy runs one video forward pass and one action-head $10$-step Euler integration per replan. Language guidance adds a null-text video pass, plan guidance adds an extended-horizon video pass, and combined guidance adds both. The video forward-pass count per replan is therefore $1$ for unguided, $2$ for language-only or plan-only guidance, and $3$ for combined guidance; the action-head cost is unchanged across all four configurations.

All videos are included in supplementary and \url{https://umishra.me/temporal-ratio/}.

\end{document}